%% file: main.tex
\pdfoutput=1
\RequirePackage[T1]{fontenc}
\documentclass[12pt]{article}
\usepackage[skip=1em plus 0.2em minus 0.1em]{parskip}
\usepackage[height=8.85in,width=6.45in]{geometry}

\usepackage[utf8]{inputenc}
\usepackage{amsmath}
\usepackage{amssymb}
\usepackage{mathtools}
\numberwithin{equation}{section}
\usepackage{slashed}
\usepackage{braket}
\usepackage{enumitem}
\usepackage[svgnames]{xcolor}
\usepackage[colorlinks,citecolor=DarkGreen,linkcolor=FireBrick,urlcolor=FireBrick,linktocpage,unicode]{hyperref}

\usepackage{tocloft}

\usepackage[hang,flushmargin,bottom]{footmisc}
\input{macros.tex}

\newcommand*{\email}[1]{\footnote{\href{mailto:#1}{\texttt{#1}}}}

\setlist[itemize,enumerate]{
  parsep=\parskip,                                   %
  itemsep=\dimexpr .3em - \parskip\relax plus 2pt,   %
  topsep=\dimexpr 6pt - \parskip\relax plus 1pt minus 1pt,
  partopsep=0pt,
  listparindent=\parindent
}

\begin{document}
\begin{titlepage}

\begin{flushright}
Last Update: Novermber 7, 2025
\end{flushright}

\vskip 2.5em
\begin{center}

{
\LARGE \bfseries %
\begin{spacing}{1.15} %
\input{title} %
\end{spacing}
}

\vskip 1em
Maojiang Su$^{\dagger*}$\email{smj@u.northwestern.edu}
\quad
Jerry Yao-Chieh Hu$^{\dagger*}$\email{jhu@u.northwestern.edu}
\quad
Sophia Pi$^{\dagger}$\email{sophiapi2026.1@u.northwestern.edu}
\quad
Han Liu$^{\dagger\ddag}$\email{hanliu@northwestern.edu}

\def\thefootnote{*}
\footnotetext{These authors contributed equally to this work.
Code is available at \url{https://github.com/MAGICS-LAB/flow_kl}.}

\vskip 1em

{\small
\begin{tabular}{ll}
 $^\dagger\;$Center for Foundation Models and Generative AI, Northwestern University, Evanston, IL 60208, USA\\
 \hphantom{$^\ddag\;$}Department of Computer Science, Northwestern University, Evanston, IL 60208, USA\\
 $^\ddag\;$Department of Statistics and Data Science, Northwestern University, Evanston, IL 60208, USA
\end{tabular}}

\end{center}

\noindent
\input{0abstract}

\end{titlepage}

{
\setlength{\parskip}{0em}
\setcounter{tocdepth}{2}
\tableofcontents
}
\setcounter{footnote}{0}

\section{Introduction}
\label{sec:intro}
\input{1intro}

\section{Preliminaries}
\label{sec:preliminary}
\input{preliminary}

\section{Kullback-Leibler (KL)  Error Bound for Flow Matching}
\label{sec:method}

\input{2method}

\section{Numerical Studies}
\label{sec:exp}\input{3exp}

\section{Discussion and Conclusion}
\label{sec:conclusion}
\input{4conclusion}

\section*{Impact Statement}
\input{impact}

\section*{Acknowledgments}
\input{x_acknowledgments}

\newpage
\appendix
\label{sec:append}
\part*{Appendix}
{
\setlength{\parskip}{-0em}
\startcontents[sections]
\printcontents[sections]{ }{1}{}
}

\input{appendix}

\clearpage
\def\arxivfont{\rm}
\bibliographystyle{plainnat}

\bibliography{refs}

\end{document}

%% file: macros.tex
\allowdisplaybreaks
\usepackage{graphicx}
\usepackage{tikz}
\usepackage{tikz-cd}
\usepackage{times}
 
\usepackage{bm}
\usepackage{physics}
\usepackage{xcolor}
\usepackage{natbib}
\usepackage{mdframed}
\usepackage{nicefrac}
\usepackage{booktabs}
\usepackage{lipsum}
\usepackage{titlesec}
\usepackage{wrapfig,lipsum,booktabs}
\usepackage{authblk}
\usepackage{blindtext}

\usepackage{algorithm}
\usepackage{algpseudocode}
\newcommand{\commentsymbol}{//}%
\algrenewcommand\algorithmiccomment[1]{\hfill {\footnotesize \commentsymbol{} #1}}

\usepackage{titletoc}
\usepackage{authblk}
\usepackage{setspace}
\usepackage{dsfont} %

\usepackage[nameinlink,capitalize,noabbrev]{cleveref}

\usepackage{CJK}

\usepackage{array}
\usepackage{bbm}
\usepackage{makecell}

\usepackage{dashbox}
\usepackage{xcolor}
\usepackage{colortbl}
\definecolor{lightyellow}{rgb}{1.0, 0.95, 0.7}
\definecolor{Blue}{rgb}{0, 0, 0.8}
\definecolor{blue}{rgb}{0,0,1}
\definecolor{darkgreen}{rgb}{0,0.40,0}
\definecolor{firebrick}{rgb}{0.698,0.133,0.133}

\definecolor{colorA}{rgb}{1,0,0}
\definecolor{colorB}{rgb}{0,0.3,1}
\definecolor{colorC}{rgb}{0.9,0.8,0.2}
\definecolor{colorD}{rgb}{0,0.65,0}
\definecolor{lesslightgray}{rgb}{0.5,0.5,0.5}
\definecolor{light-gray}{gray}{0.95}

\let\tilde\widetilde
\let\hat\widehat

\newcommand{\calH}{\mathcal{H}}

\newcommand{\calP}{\mathcal{P}}

\newcommand{\calR}{\mathcal{R}}

\def\R{\mathbb{R}}

\DeclareMathOperator*{\E}{{\mathbb{E}}} %

\let\cite\citep 

\usepackage{amsthm}
\usepackage[many]{tcolorbox}
\usepackage{etoolbox}
\BeforeBeginEnvironment{proof}{\par\vspace{-1\baselineskip}}
\AfterEndEnvironment  {proof}{\par\vspace{-1.5\baselineskip}}

\newtheoremstyle{theoremstyle}
  {.5\baselineskip} %
  {.5\baselineskip} %
  {}                  %
  {}                  %
  {\bfseries}        %
  {.}                 %
  {1em}               %
  {}                  %

\theoremstyle{theoremstyle}
\newtheorem{theorem}{Theorem}[section]
\newtheorem{lemma}{Lemma}[section]

\newtheorem{proposition}{Proposition}[section]
\newtheorem{definition}{Definition}[section]
\newtheorem{assumption}{Assumption}[section]
\newtheorem{remark}{Remark}[section]

\tcolorboxenvironment{theorem}{
  breakable,
  colback=black!10,
  colframe=white,%
  width=\dimexpr\linewidth+10pt\relax,%
  enlarge left by=-5pt,%
  enlarge right by=-5pt,%
  boxsep=5pt,%
  boxrule=0pt,
  left=0pt,right=0pt,top=0pt,bottom=0pt,
  sharp corners,
  before skip=0.5\baselineskip, %
  after skip=0.5\baselineskip,  %
  fonttitle=\bfseries, %
  coltitle=black %
}
\tcolorboxenvironment{remark}{
  blanker,
  breakable,
  before skip=.8\baselineskip,  %
  after  skip=.8\baselineskip   %
}

\tcolorboxenvironment{proposition}{
  breakable,
  colback=black!10,
  colframe=white,%
  width=\dimexpr\linewidth+10pt\relax,%
  enlarge left by=-5pt,%
  enlarge right by=-5pt,%
  boxsep=5pt,%
  boxrule=0pt,
  left=0pt,right=0pt,top=0pt,bottom=0pt,
  sharp corners,
  before skip=0.5\baselineskip, %
  after skip=0.5\baselineskip,  %
  fonttitle=\bfseries, %
  coltitle=black %
}

\tcolorboxenvironment{lemma}{
  breakable,
  colback=black!10,
  colframe=white,%
  width=\dimexpr\linewidth+10pt\relax,%
  enlarge left by=-5pt,%
  enlarge right by=-5pt,%
  boxsep=5pt,%
  boxrule=0pt,
  left=0pt,right=0pt,top=0pt,bottom=0pt,
  sharp corners,
  before skip=0.5\baselineskip, %
  after skip=0.5\baselineskip,  %
  fonttitle=\bfseries, %
  coltitle=black %
}

\tcolorboxenvironment{corollary}{
  breakable,
  colback=black!10,
  colframe=white,%
  width=\dimexpr\linewidth+10pt\relax,%
  enlarge left by=-5pt,%
  enlarge right by=-5pt,%
  boxsep=5pt,%
  boxrule=0pt,
  left=0pt,right=0pt,top=0pt,bottom=0pt,
  sharp corners,
  before skip=0.5\baselineskip, %
  after skip=0.5\baselineskip,  %
  fonttitle=\bfseries, %
  coltitle=black %
}

\tcolorboxenvironment{definition}{
  breakable,
  colback=black!10,
  colframe=white,%
  width=\dimexpr\linewidth+10pt\relax,%
  enlarge left by=-5pt,%
  enlarge right by=-5pt,%
  boxsep=5pt,%
  boxrule=0pt,
  left=0pt,right=0pt,top=0pt,bottom=0pt,
  sharp corners,
  before skip=0.5\baselineskip, %
  after skip=0.5\baselineskip,  %
  fonttitle=\bfseries, %
  coltitle=black %
}
\tcolorboxenvironment{assumption}{
  breakable,
  colback=black!10,
  colframe=white,%
  width=\dimexpr\linewidth+10pt\relax,%
  enlarge left by=-5pt,%
  enlarge right by=-5pt,%
  boxsep=5pt,%
  boxrule=0pt,
  left=0pt,right=0pt,top=0pt,bottom=0pt,
  sharp corners,
  before skip=0.5\baselineskip, %
  after skip=0.5\baselineskip,  %
  fonttitle=\bfseries, %
  coltitle=black %
}

\crefname{theorem}{Theorem}{Theorems}
\crefname{proposition}{Proposition}{Propositions}
\crefname{lemma}{Lemma}{Lemmas}
\crefname{corollary}{Corollary}{Corollaries}
\crefname{definition}{Definition}{Definitions}
\crefname{assumption}{Assumption}{Assumptions}
\crefname{remark}{Remark}{Remarks}
\crefname{problem}{Problem}{Problems}
\crefname{property}{Property}{property}

\tcolorboxenvironment{hypothesis}{
  breakable,
  colback=black!10,
  colframe=white,%
  width=\dimexpr\linewidth+10pt\relax,%
  enlarge left by=-5pt,%
  enlarge right by=-5pt,%
  boxsep=5pt,%
  boxrule=0pt,
  left=0pt,right=0pt,top=0pt,bottom=0pt,
  sharp corners,
  before skip=0.5\baselineskip, %
  after skip=0.5\baselineskip,  %
  fonttitle=\bfseries, %
  coltitle=black %
}
\crefname{hypothesis}{Hypothesis}{Hypothesises}

\tcolorboxenvironment{fact}{
  breakable,
  colback=black!10,
  colframe=white,%
  width=\dimexpr\linewidth+10pt\relax,%
  enlarge left by=-5pt,%
  enlarge right by=-5pt,%
  boxsep=5pt,%
  boxrule=0pt,
  left=0pt,right=0pt,top=0pt,bottom=0pt,
  sharp corners,
  before skip=0.5\baselineskip, %
  after skip=0.5\baselineskip,  %
  fonttitle=\bfseries, %
  coltitle=black %
}
\crefname{fact}{Fact}{Facts}

\tcolorboxenvironment{example}{
  breakable,
  colback=black!10,
  colframe=white,%
  width=\dimexpr\linewidth+10pt\relax,%
  enlarge left by=-5pt,%
  enlarge right by=-5pt,%
  boxsep=5pt,%
  boxrule=0pt,
  left=0pt,right=0pt,top=0pt,bottom=0pt,
  sharp corners,
  before skip=0.5\baselineskip, %
  after skip=0.5\baselineskip,  %
  fonttitle=\bfseries, %
  coltitle=black %
}
\crefname{example}{Example}{Examples}

\tcolorboxenvironment{question}{
  breakable,
  colback=black!10,
  colframe=white,%
  width=\dimexpr\linewidth+10pt\relax,%
  enlarge left by=-5pt,%
  enlarge right by=-5pt,%
  boxsep=5pt,%
  boxrule=0pt,
  left=0pt,right=0pt,top=0pt,bottom=0pt,
  sharp corners,
  before skip=0.5\baselineskip, %
  after skip=0.5\baselineskip,  %
  fonttitle=\bfseries, %
  coltitle=black %
}

\crefname{question}{Question}{Questions}
\numberwithin{equation}{section}
\numberwithin{theorem}{section}
\numberwithin{proposition}{section}
\numberwithin{definition}{section}
\numberwithin{lemma}{section}
\numberwithin{assumption}{section}
\numberwithin{remark}{section}

\usepackage{lipsum}

\newcommand*{\annot}[1]{\tag*{\footnotesize{\textcolor{black!50}{\big(#1\big)}}}}

\makeatletter
\let\save@mathaccent\mathaccent
\newcommand*\if@single[3]{%
    \setbox0\hbox{${\mathaccent"0362{#1}}^H$}%
    \setbox2\hbox{${\mathaccent"0362{\kern0pt#1}}^H$}%
    \ifdim\ht0=\ht2 #3\else #2\fi
}
\newcommand*\rel@kern[1]{\kern#1\dimexpr\macc@kerna}
\newcommand*\widebar[1]{\@ifnextchar^{{\wide@bar{#1}{0}}}{\wide@bar{#1}{1}}}
\newcommand*\wide@bar[2]{\if@single{#1}{\wide@bar@{#1}{#2}{1}}{\wide@bar@{#1}{#2}{2}}}
\newcommand*\wide@bar@[3]{%
    \begingroup
    \def\mathaccent##1##2{%
        \let\mathaccent\save@mathaccent
        \if#32 \let\macc@nucleus\first@char \fi
        \setbox\z@\hbox{$\macc@style{\macc@nucleus}_{}$}%
        \setbox\tw@\hbox{$\macc@style{\macc@nucleus}{}_{}$}%
        \dimen@\wd\tw@
        \advance\dimen@-\wd\z@
        \divide\dimen@ 3
        \@tempdima\wd\tw@
        \advance\@tempdima-\scriptspace
        \divide\@tempdima 10
        \advance\dimen@-\@tempdima
        \ifdim\dimen@>\z@ \dimen@0pt\fi
        \rel@kern{0.6}\kern-\dimen@
        \if#31
        \overline{\rel@kern{-0.6}\kern\dimen@\macc@nucleus\rel@kern{0.4}\kern\dimen@}%
        \advance\dimen@0.4\dimexpr\macc@kerna
        \let\final@kern#2%
        \ifdim\dimen@<\z@ \let\final@kern1\fi
        \if\final@kern1 \kern-\dimen@\fi
        \else
        \overline{\rel@kern{-0.6}\kern\dimen@#1}%
        \fi
    }%
    \macc@depth\@ne
    \let\math@bgroup\@empty \let\math@egroup\macc@set@skewchar
    \mathsurround\z@ \frozen@everymath{\mathgroup\macc@group\relax}%
    \macc@set@skewchar\relax
    \let\mathaccentV\macc@nested@a
    \if#31
    \macc@nested@a\relax111{#1}%
    \else
    \def\gobble@till@marker##1\endmarker{}%
    \futurelet\first@char\gobble@till@marker#1\endmarker
    \ifcat\noexpand\first@char A\else
    \def\first@char{}%
    \fi
    \macc@nested@a\relax111{\first@char}%
    \fi
    \endgroup
    }
\makeatother

%% file: title.tex
On Flow Matching KL Divergence

%% file: 0abstract.tex
We derive a deterministic, non-asymptotic upper bound on the Kullback-Leibler (KL) divergence of the flow-matching distribution approximation.
In particular, if the $L_2$ flow-matching loss is bounded by $\epsilon^2 > 0$, then the KL divergence between the true data distribution and the estimated distribution is bounded by $A_1 \epsilon + A_2 \epsilon^2$. 
Here, the constants $A_1$ and $A_2$ depend only on the regularities of the data and velocity fields.
Consequently, this bound implies statistical convergence rates of Flow Matching Transformers under the Total Variation (TV) distance.
We show that, flow matching achieves nearly minimax-optimal efficiency in estimating smooth distributions.
Our results make the statistical efficiency of flow matching comparable to that of diffusion models under the TV distance.
Numerical studies on synthetic and learned velocities corroborate our theory.

\vfill
\textbf{Keywords:} Flow Matching, Probability Flow ODE, Non-Asymptotic Convergence, Kullback-Leibler (KL) Divergence Error Bounds, Flow Matching Transformer

%% file: 1intro.tex
We establish a rigorous upper bound on the Kullback-Leibler (KL) divergence between the true data distribution and the distribution estimated by flow matching, expressed in terms of the $L_2$ flow matching training loss.
Furthermore, we prove Flow Matching Transformers achieves almost minimax optimal
convergence rate under the Total Variation (TV) metric.
A theoretical understanding is crucial in the current era of rapidly advancing generative AI.
In particular, Flow Matching (FM) \cite{lipman2022flow, liu2022flow, albergo2022building} is a central paradigm for training continuous-time generative models.
Instead of maximizing likelihoods, FM directly learns a deterministic velocity field that transports a simple base distribution (e.g., Gaussian) to a complex data distribution through the continuity equation.
Its simple training objective and efficient inference enable state-of-the-art performance across diverse domains, including image generation \citep{esser2024scaling}, speech synthesis \citep{le2023voicebox}, video generation \citep{polyak2025moviegencastmedia}, and robotics \citep{black2410pi0}.

Despite its empirical success, the theoretical understanding of flow matching remains limited.
Recent theoretical studies \citep{chen2022sampling,gentiloni2024theoretical,block2020generative,de2022convergence} investigate the distributional approximation error of diffusion-based generative models under stochastic differential equation (SDE) sampling.
For flow-based generative models, \citet{albergo2022building,benton2023error} analyze the distributional error of flow matching under the 2-Wasserstein distance. \citet{albergo2023stochastic} derive KL-based bounds for stochastic flows by injecting a small diffusion term into the sampling process.
Although these results provide valuable insights, they primarily focus on stochastic sampling or Wasserstein-type metrics, which capture geometric transport cost rather than information-theoretic discrepancy.
More recently, \citet{fukumizu2024flow,jiao2024convergence} investigate the convergence rates of flow matching under the 2-Wasserstein distance.
However, the statistical convergence rate under more interpretable and task-relevant metrics, such as the Total Variation (TV) distance, is unexplored.

In contrast, our work develops a pure ODE-based theory for deterministic flow matching and establishes an explicit Kullback-Leibler (KL) divergence bound between the true and learned distributions.
The KL metric is more relevant for likelihood-based generative modeling \cite{schulman2015trust,abdolmaleki2018maximum,haarnoja2018soft,shao2024deepseekmath}, as it quantifies information loss and statistical discrepancy rather than geometric displacement.
Building on this result, we further derive statistical convergence rates for Flow Matching Transformers under the Total Variation (TV) metric.
The KL bound provides a direct and information-theoretic control of distributional error, allowing us to obtain TV convergence in a more explicit and interpretable way than studies based on the 2-Wasserstein distance.
Together, our KL error bounds and convergence rates under TV distance complement existing Wasserstein-based studies, forming a unified foundation for understanding flow matching.

\textbf{Contributions.} Our contributions are two-fold:
\begin{itemize}
    \item \textbf{Kullback-Leibler Error Bounds for Flow Matching.}
    We establish a  \textit{deterministic, non-asymptotic upper bound} on the Kullback-Leibler (KL) divergence between the true data distribution $p_1$ and the flow-matching-estimated distribution $q_1$, expressed in terms of the $L_2$ flow matching training loss.
    Following \citet{albergo2023stochastic}, we introduce the KL Evolution Identity for Continuity Flows (\cref{lem:kl_identity}) and apply Grönwall’s inequality (\cref{lem:Gronwall's}) to control the temporal evolution of the divergence.
    Under mild smoothness and regularity assumptions \cref{ass:fm_loss_bound,ass:regu_score,ass:regu_velocity}, \cref{thm:kl_bound} shows that $\text{KL}(p_1||q_1) \leq A_1\epsilon + A_2\epsilon^2$, where 
    $A_1,A_2$ are regularity and smoothness constants and $\epsilon^2$ denotes the $L_2$ flow matching training loss.
    This result provides the first information-theoretic guarantee that links training loss to distributional approximation accuracy.
    \item \textbf{Convergence Rates under Total Variation Distance.}
    Building on the KL bound, we derive statistical convergence rates for Flow Matching Transformers under the \textit{Total Variation (TV)} metric.
    We begin by introducing the \textit{Hölder space} (\cref{def:holder_norm_space}).
    Under the Hölder-smoothness assumption (\cref{assumption:density_function_assumption_1}),
    we establish the convergence rate stated in \cref{thm:dist_esti_tv},
    and further show that flow matching achieves \textit{nearly minimax-optimal convergence} under specific regularity conditions (\cref{thm:minimax}).
    This result complements prior analyses based on the 2-Wasserstein distance by providing a more direct and interpretable characterization of the probabilistic approximation ability of flow matching.
\end{itemize}

\textbf{Organization.}
\cref{sec:preliminary} reviews the preliminary concepts of standard flow matching.
\cref{sec:method} presents the KL error bound for the distributional approximation error of flow matching.
\cref{sec:rates} establishes the statistical convergence rates of Flow Matching Transformers and proves that flow matching achieves nearly minimax-optimal convergence.
\cref{sec:exp} reports the empirical results that validate our theoretical analysis.
\cref{sec:conclusion} concludes the paper and discusses the implications of our findings.
The appendix provides related works (\cref{sec:related_work}), and detailed proofs of the main results (\cref{sec:proofs}).

\textbf{Notation.}
We denote the index set $\{1,\ldots,I\}$ by $[I]$.
Let $x[i]$ denote the $i$-th component of a vector $x$.
Let $\mathbb{Z}$ denote integers and $\mathbb{Z}_{+}$ denote positive integers.
Given random variables $X$ and $Y$ with marginal densities $\mu_x$ and $\mu_y$ respectively, we denote the 2-Wasserstein distance between $\mu_x$ and $\mu_y$ by $W_2(\mu_x, \mu_y)$.
Given a matrix $Z \in \R^{d \times L}$, $\| Z \|_2$ and $\| Z \|_{\rm F}$ denote the $2$-norm and the Frobenius norm.
Let $u^k \in \R^d$ be column vectors for $k \in [K]$, we denote $\text{col}(u^1,\ldots,u^K) \in \R^{kd}$ as the vertical concatenation of $u^1, \ldots, u^K$.
Let $\text{Div}~\cdot $ be the divergence operator.

%% file: preliminary.tex
In this section, we provide an overview of the flow matching generative modeling following \cite{lipman2024flow} and \cite[Section 2]{su2025high}.

\subsection{Flow Matching}

\textbf{Flow Model.}
\label{sec:flow_model}
The flow model transforms $X_0 = x_0$ drawn from a source distribution $P$ (e.g., a Gaussian) into samples $X_1 = x_1$ from a target distribution $Q$.
A flow $\psi: [0,1] \times \R^d \to \R^d$ is a time-dependent mapping $\psi:(t,x) \mapsto \psi_t(x)$ that evolves the input $x$ over time.
The flow model is a continuous-time Markov process $(X_t)_{0\leq t \leq 1}$ defined by applying a flow $\psi_t$ to the random variable $X_0 \sim P$:
\begin{align*}
    X_t = \psi_t(X_0),\quad t \in [0,1].
\end{align*}
On the other hand, a time-dependent velocity field $u: \R^d \times [0,1] \rightarrow \R^d$ implementing $u:(x,t) \mapsto u(\cdot,t)$ defines a unique flow $\psi$ via the following ordinary differential equation (ODE): 
\begin{align}
\label{eqn:flow_ODE}
    \dv{\psi_t}{t} = u(\psi_t(x),t) \quad\text{with initial condition}\quad
    \psi_0(x) = x.
\end{align}
Given a flow $\psi_t$, the marginal probability density function (PDF) of flow model $X_t = \psi_t(X_0) \sim p_t$ is a continuous-time  probability path $(p_t)_{0 \leq t \leq 1}$. 
The evolution of marginal probability densities $p_t$ follows the push-forward equation:
\begin{align}
\label{eqn:flow_to_prob}
p_t(x) = [\psi_t]_* p_0(x) 
:= p_0(\psi^{-1}_t(x)) \cdot
\abs{\det[\pdv{\psi^{-1}_t}{x}(x)]}.
\end{align}
By the equivalence between flows and velocity fields \cite{lipman2024flow}, any invertible $C^1$ diffeomorphism $\psi_t$ induces a unique smooth conditional velocity field $u(x,t)$ given by
\begin{align}
\label{eqn:flow_to_velocity}
    u(x,t) = \dot{\psi_t} (\psi_t^{-1}(x)), \quad\text{with}\quad
    \dot{\psi_t} = \dv{}{t} \psi_t.
\end{align}
Equations \eqref{eqn:flow_to_prob} and \eqref{eqn:flow_to_velocity} characterize the relationship among the probability path $p_t$, the flow $\psi_t$, and the velocity field $u$.
The continuity equation provides a direct link between $u$ and $p_t$:
\begin{align}
\label{eq:continuity_eq_pre}
    \pdv{}{t} p_t(x)
    + \div (p_t(x) u(x,t)) =0.
\end{align}
For an arbitrary probability path $p_t$, we define a velocity field $u(x,t)$ that \textit{generates} $p_t$ if its flow $\psi_t$ satisfies \eqref{eqn:flow_ODE}. 
The mass conservation theorem \cite{lipman2024flow,villani2008optimal} ensures consistency between the continuity equation and flow ODE \eqref{eqn:flow_ODE}:
a pair $(u,p_t)$ satisfying \eqref{eq:continuity_eq_pre} for $t \in [0,1]$ corresponds to a field velocity $u(x,t)$ that generates $p_t$.

Continuous Normalizing Flow \cite{chen2018cnf} models the velocity field $u(x,t)$ with a neural network $u_\theta$.
Once we obtain a well-trained $u_\theta$, we generate samples from solving ODE \eqref{eqn:flow_ODE}.

\textbf{Flow Matching.}
Instead of training flow model by maximizing the log-likelihood of training data \cite{chen2018cnf}, flow matching \cite{lipman2022flow} is a simulation-free framework to train flow generative models
without the need of solving ODEs during training.
The Flow Matching objective is designed to match the probability path $(p_t)_{0 \leq t \leq 1}$, which allows us to flow from source $p_0 = P$ to target $p_1 = Q$.
Suppose $u$ generates such probability path $p_t$, the flow matching loss is 
\begin{align}
\label{eqn:fm_loss_1}
    \mathcal{L}_{\text{FM}}(\theta) = \E_{t,X_t \sim p_t}[\|u_\theta(X_t,t)-u(X_t,t)\|_2^2], 
\end{align}
where $t \sim U[0,1]$, $u_\theta(x,t)$ is a neural network with parameter $\theta$. 
Flow Matching simplifies the problem of designing a probability path $p_t$ and its
corresponding velocity field $u(x,t)$ by adopting a conditional strategy.
Formally, conditioning on any arbitrary random vector $Z \in \R^m$ with PDF $p_Z$, the marginal probability path $p_t$ satisfies
\begin{align}
\label{eqn:marginal_p}
    p_t(x) 
    = \int p_t(x|z) p_Z(z) \dd z.
\end{align}
Suppose conditional velocity field $u(x|z,t)$ generates $p_{t}(x|z)$, \citet{lipman2022flow} show that following marginal velocity field $u(x,t)$ generates marginal probability path $p_t$ under mild assumptions:
\begin{align}
\label{eqn:marginal_u}
    u(x,t) 
    := \int u(x|z,t) p_{Z|t}
    (z|x) \dd z \quad \text{with}\quad
    p_{Z|t}(z|x) 
    = \frac{p_t(x|z) p_Z(z)}{p_t(x)},
\end{align}
where the second equation follows from the Bayes' rule.
Combining above, the tractable conditional flow matching loss $\mathcal{L}_{\text{CFM}}$, which satisfies $\nabla_{\theta} \mathcal{L}_{\text{CFM}}(\theta) = \nabla_{\theta} \mathcal{L}_{\text{FM}} (\theta)$, is defined as:
\begin{align}
\label{eqn:con_fm_loss}
    \mathcal{L}_{\text{CFM}}(\theta) = \E_{t, Z \sim p_Z, X_t \sim p_{t}(\cdot|Z)}[\|u_\theta(X_t,t)-u(X_t|Z,t)\|_2^2].
\end{align}
\textbf{Affine Conditional Flows.}
\label{sec:affine_condition}
The conditional flow matching loss works with any choice of conditional probability path
and conditional velocity fields.
In this paper, we consider the affine conditional flow with independent data coupling following \cite{lipman2022flow,lipman2024flow}:
\begin{align}
\label{eqn:aff_con_flow}
    \psi_t(x|x_1) 
    = \mu_t x_1 + \sigma_t x,
\end{align}
where $\mu_t,\sigma_t:[0,1] \rightarrow [0,1]$ are monotone smooth functions satisfying
\begin{align}
\label{eqn:aff_con_flow_coeff}
    \mu_0 = \sigma_1 = 0, ~
    \mu_1 = \sigma_0 = 1, \quad
    \text{and} \quad 
    \dv{\mu_t}{t}, -\dv{\sigma_t}{t} > 0 \quad
    \text{for} \quad t \in (0,1).
\end{align}
Setting $Z=X_1 \sim Q$, $X_0 \sim N(0,I)$, the flow $\psi_t$ induces the probability flow 
$p_t(X_t|X_1) = N(\mu_t X_1, \sigma_t^2 I)$ and velocity field 
\begin{align}
\label{eqn:velocity_con}
    u(x|x_1,t) 
    = \dot{\psi_t} (\psi_t^{-1}(x|x_1)|x_1)
    = \frac{\dot{\sigma}_t(x -\mu_t x_1)}{\sigma_t} + \dot{\mu}_t x_1.
\end{align}
Further, using the law of unconscious statistician with $X_t = \psi_t(X_0|X_1)$, the conditional flow matching loss takes the form 
\begin{align}
    \mathcal{L}_{\text{CFM}}(\theta) 
    = \E_{t, X_1 \sim q, X_0 \sim N(0,I)}\big[ \|u_\theta(\mu_t X_1 + \sigma_t X_0, t)-(\dot{\mu_t}X_1 +\dot{\sigma_t}X_0)\|_2^2 \big].
    \label{eqn:con_fm_loss_2}
\end{align}
In practice, for collected i.i.d. data points $\{x_i\}_{i=1}^n$, \eqref{eqn:con_fm_loss_2} is implemented with Monte-Carlo simulation.
To avoid instability,
we often clip the interval $[0,1]$ with $t_0$ and $T$.
Namely, for any velocity estimator $u^\theta$, we consider the empirical loss function $\hat{\mathcal{L}}_{\text{CFM}}(\theta)$:
\begin{align}
\label{eqn:con_fm_loss_em}
    \hat{\mathcal{L}}
    _{\text{CFM}}(\theta)
    := \frac{1}{n} \sum_{i=1}^n
    \int_{t_0}^T \frac{1}{T-t_0}
    \E_{X_0 \sim N(0,I)} \big[\|u_\theta(\mu_t x_i + \sigma_t X_0, t)-(\dot{\mu_t}x_i +\dot{\sigma_t}X_0)\|_2^2 \big] \dd t.
\end{align}

%% file: 2method.tex
In this section, we provide a novel upper bound on the Kullback-Leibler (KL) divergence between the true data distribution and the Flow Matching (FM) estimated distribution.
Our analysis starts from a fundamental KL evolution identity (\cref{lem:kl_identity}) that characterizes how the divergence evolves along the probability flow \cite{albergo2023stochastic}.
We then show that, under mild regularity assumptions on the velocity fields and density paths, this identity yields an explicit upper bound on the terminal KL divergence in terms of the $L_2$ Flow Matching training loss (\cref{thm:kl_bound}).

\citet{albergo2023stochastic} presents a \textit{evolution identity} that characterizes the evolution of the KL divergence between the true distribution and the flow-matching estimated distribution.

\begin{lemma}[KL Evolution Identity for Continuity Flows; Lemma 21 of \cite{albergo2023stochastic}]
\label{lem:kl_identity}
    Let two velocities field $u(x,t),v(x,t) \in C([0,1];(C^1(\R^d))^d)$.
    Let $p_t$ and $q_t$ be two paths of differentiable probability densities on $\R^d$ evolving under the continuity equations 
    \begin{align}
    \label{eqn:cont_eqn_main_text}
        \pdv{p_t(x)}{t} + \div (p_t(x) u(x, t)) = 0, \quad
        \pdv{q_t(x)}{t} + \div (q_t(x) v(x, t)) = 0,
    \end{align}
    with same initial distribution $p_0 = q_0$. 
    Then for all $t \in [0,1]$,
    \begin{align}
    \dv{}{t} \text{KL}(p_t||q_t)
    = \E_{x \sim p_t}[ \left(u(x,t)-v(x,t) \right)^\top 
    \left( \nabla \log p_t(x) - \nabla \log q_t(x) \right)]. 
    \label{eq:evo_identity}
    \end{align}
\end{lemma}
\begin{proof}
For completeness, we restate the proof in  \cref{sec:proof_kl_identity}.
\end{proof}
\cref{lem:kl_identity} shows that the time derivative of the KL divergence between the true path $p_t$ and the estimated path $q_t$ equals the expected inner product between the velocity error and the Stein score error of the two flows.
Consequently, the terminal KL divergence is the path integral of the multiplication of velocity discrepancy and stein score discrepancy along the trajectory.
\begin{lemma}
[KL Difference for Continuity Flows]
\label{coro:kl_terminal_identity}
    Let two velocities field $u(x,t),v(x,t) \in C([0,1];(C^1(\R^d))^d)$.
    Let $p_t$ and $q_t$ be two paths of differentiable probability densities on $\R^d$ evolving under the continuity equations 
    \begin{align*}
        \pdv{p_t(x)}{t} + \div (p_t(x) u(x, t)) = 0, \quad
        \pdv{q_t(x)}{t} + \div (q_t(x) v(x, t)) = 0,
    \end{align*}
    with same initial distribution $p_0 = q_0$. 
    Then for all $t \in [0,1]$,
    \begin{align}
    \label{eq:kl_ter_id}
    \text{KL}(p_1||q_1)
    = & ~ \int_0^1 \E_{x \sim p_t}[(u(x,t)-v(x,t))^\top (\nabla \log p_t(x) - \nabla \log q_t(x)] \dd t.   
    \end{align}
\end{lemma}
\begin{proof}
    Taking the integral of both sides of \eqref{eq:evo_identity} with respect to time $t$.
\end{proof}

By applying the KL Difference for Continuity Flows (\cref{coro:kl_terminal_identity}), we bound terminal the KL error for flow matching $\text{KL}(p_1||q_1)$ in terms of the flow-matching loss and structural regularity of the underlying flows.
To control the stein score error, we introduce additional conditions on training error and Lipschitz smoothness of velocities and data score.

\begin{assumption}[Bound on $L_2$ Flow Matching Loss]
\label{ass:fm_loss_bound}
Let $p_t$ denote the probability path induced by $u_t$.
Assume the true velocity $u(x,t)$ and its estimators $v(x,t)$ satisfy
\begin{align*}
    \mathcal{L}_{\text{FM}} 
    = \E_{t,x \sim p_t}[\|u(x,t)-v(x,t)\|_2^2] \leq \epsilon^2.
\end{align*}
For each $t \in [0,1]$, define the point-wise error $\epsilon(t)$ as
\begin{align*}
    \epsilon(t) := \sqrt{\E_{x \sim p_t}\!\left[\|u(x,t) - v(x,t)\|_2^2\right]}.
\end{align*}
\end{assumption}

\begin{assumption}[Regularity of Data Path]
\label{ass:regu_score}
    Assume the true probability path $p_t$ satisfy that for any $t\in[0,1]$ and $x \in \R^d$, 
    \begin{align*}
        \| \nabla \log p_t(x) \|_2 \leq B_p(t), \quad
        \| \nabla(\nabla \log p_t(x)) \|_2 \leq U_p(t).
    \end{align*}
\end{assumption}

\begin{assumption}[Regularity of  Velocity Fields]
\label{ass:regu_velocity}
Assume velocities $u(x,t),v(x,t) \in C^2$ satisfy for any $t\in[0,1]$ and $x \in \R^d$,
$\abs{\div u - \div v}\leq K(t)$ and following hold
\begin{align*}
    \|u(x,t)\|_{2} 
    \leq & M(t), \quad
    \|v(x,t)\|_{2} 
    \leq M(t), \\
    \|\nabla u(x,t)\|_{2} 
    \leq & L(t), \quad
    \|\nabla v(x,t)\|_{2} 
    \leq L(t), \\
     \|\nabla(\div u(x,t))\|_{2} 
     \leq & H(t), \quad
     \|\nabla(\div v(x,t))\|_{2} \leq H(t),
\end{align*}
\end{assumption}

\cref{ass:fm_loss_bound} is a natural assumption on the training error.
\cref{ass:regu_score} and \cref{ass:regu_velocity} is essential for applying the Cauchy-Schwarz inequality to expectations and ensuring integrability.

Before we demonstrate our main result, we introduce the Gr\"{o}nwall's Inequality as a key helping lemma.
It provides a way to control the score error through its derivative upper bound.
\begin{lemma}
[Gr\"{o}nwall's Inequality; \cite{gronwall1919note}]
\label{lem:Gronwall's}
Let $a,b\in\R$ with $a < b$.
Suppose functions $g(t),y(t),h(t) \in C^1[a,b]$.
Then,
if $y(t)$ is differentiable on $[a,b]$ and satisfies:
\begin{align*}
    \dv{}{t}y(t) \leq y(t)g(t) + h(t), \quad t \in [a,b].
\end{align*}
Define $\Gamma(t) = \int_a^t g(s) \dd s$, then it holds
\begin{align*}
    y(t) 
    \leq y(a) e^{\Gamma(t)} + \int_a^t e^{\Gamma(t) - \Gamma(\tau)} h(\tau) \dd \tau.
\end{align*}
\end{lemma}

We now bound the terminal KL divergence using the flow-matching loss and the regularity constants defined in \cref{ass:fm_loss_bound}, \cref{ass:regu_score} and \cref{ass:regu_velocity}.
\begin{theorem}[Flow Matching KL Error Bounds]
\label{thm:kl_bound}
Assume \cref{ass:fm_loss_bound,ass:regu_score,ass:regu_velocity} hold. 
If the initial distributions $p_0=q_0$, then 
\begin{align*}
    \text{KL}(p_1||q_1)
    \leq & ~ \epsilon ~ 
    \sqrt{\int_0^1 \E_{x \sim p_t}
    [\|\nabla \log p_t(x) - \nabla \log q_t(x)\|_2^2] \dd t}. 
\end{align*}

Let the lipschitz constants $U_p(t),B_p(t),K(t),L(t),M(t),H(t)$ be as defined in \cref{ass:regu_score} and \cref{ass:regu_velocity}, 
then we bound the terminal KL divergence by 
\begin{align*}
    \text{KL}(p_1||q_1)
    \leq A_1 \epsilon + A_2 \epsilon^2,
\end{align*} 
where
\begin{align*}
    A_1:= & ~ \exp{\int_0^1 L(t) + K(t) + B_p(t) M(t)\dd t} \cdot \int_0^1 2L(t)B_p(t) + 2H(t)\dd t, \\
    A_2:= & ~ \exp{\int_0^1 L(t) + K(t) + B_p(t) M(t)\dd t} \cdot \sqrt{\int_0^1 U_p(t)^2 \dd t}.
\end{align*}
\end{theorem}
\begin{proof}[Proof Sketch]
Please see \cref{sec:proof_kl_bound} for a detailed proof.
\end{proof}

\cref{thm:kl_bound} implies that when the velocity estimator achieves a small flow-matching loss 
$\epsilon$, the terminal KL divergence is guaranteed to remain small, with constants determined by the Lipschitz and regularity properties of the underlying flows.
As $\epsilon \to 0$, the KL divergence exhibits an asymptotically linear convergence rate with respect to 
$\epsilon$.
This result establishes the first explicit asymptotic relation between flow matching training error and KL divergence.
This provides a theoretical justification for the empirical stability of flow matching generative models.

\section{Convergence Rates of Flow Matching in Total Variation Distance}
\label{sec:rates}
In this section, we establish the statistical convergence rates of Flow Matching Transformers for large sample size under the Total Variation (TV) distance.
Previous studies \citep{fukumizu2024flow,jiao2024convergence,su2025high} analyze the convergence properties of Flow Matching models under the $2$-Wasserstein distance.
By exploiting the KL error bounds (\cref{thm:kl_bound}), we extend these results to the TV distance, which provides a more direct characterization of the discrepancy between probability distributions.
In \cref{thm:dist_esti_tv}, we present the convergence rate of Flow Matching Transformers under a general Hölder-smoothness assumption on the data distribution (\cref{assumption:density_function_assumption_1}).
Furthermore, in \cref{thm:minimax}, we show that Flow Matching Transformers achieve an almost minimax-optimal convergence rate under specific regularity conditions.

Estimating probability distributions whose densities lie within a Hölder ball is a common setting in nonparametric statistics and statistical rates studies \citep{gyorfi2002distribution,takezawa2005introduction,fu2024unveil,su2025theoretical}, as Hölder functions possess well-understood smoothness and approximation properties.
We begin by introducing the formal definition of the Hölder space.
\begin{definition}[H\"{o}lder Space]
\label{def:holder_norm_space}
Let $\alpha \in \mathbb{Z}_{+}^{d}$, and let $\beta = {k_1} + \gamma$ denote the smoothness parameter, where ${k_1} = \lfloor \beta \rfloor$ and $\gamma \in [0,1)$. 
Given a function $f: \R^{d} \rightarrow \R$, the H\"{o}lder space $\calH^{\beta}(\R^{d})$ is defined as the set of $\alpha$-differentiable functions satisfying:
$\calH^{\beta}(\R^{d}) \coloneqq \left\{ f: \R^{d} \rightarrow \R \mid \|f\|_{\calH^{\beta}(\R^{d})} < \infty \right\}$,
where the H\"{o}lder norm $\| f \|_{\calH^{\beta}(\R^{d})}$ satisfies:
\begin{align*}
\| f \|_{\calH^{\beta}(\R^{d})} \coloneqq \sum_{\| \alpha \|_1 < {k_1}} \sup_{x} \left| \partial^{\alpha} f(x) \right| + \max_{\alpha : \|\alpha\|_{1} = {k_1}} \sup_{x \neq x'} \frac{\left| \partial^{\alpha} f(x) - \partial^{\alpha} f(x') \right|}{\|x - x'\|_{\infty}^{\gamma}}.
\end{align*}
Also, we define the H\"{o}lder ball of radius $B$ by
$
\calH^{\beta}(\R^{d}, B) \coloneqq \left\{ f : \R^{d} \rightarrow \R \mid \| f \|_{\calH^{\beta}(\R^{d})} < B \right\}.$
\end{definition}
We assume the data distribution $p_1(x)$ is H\"{o}lder-smooth and light-tail following.
\begin{assumption}[Generic H\"{o}lder Smooth Data]
\label{assumption:density_function_assumption_1}
The density function $p_1(x)$ belongs to H\"{o}lder ball of radius $B>0$ with H\"{o}lder index $\beta>0$ (\cref{def:holder_norm_space}),
denoted by $p_1(x)\in\calH^{\beta}(\R^{d_x},B)$.
Also, 
there exist constant $C_{1}, C_{2}>0$ such that $p_1(x)\leq C_{1}\exp(-C_{2}\norm{x_1}_{2}^{2}/2)$.
\end{assumption}

For $t_0,T \in [0,1]$, we evaluate the performance of estimator $u_\theta$ through risk function $\mathcal{R}(u_\theta)$:
\begin{align*}
    \mathcal{R}(u_\theta)
    := 
    \frac{1}{T-t_0} \int_{t_0}^T \E_{x_t \sim p_t} \big[ \| u(x, t) - u_\theta( x, t ) \|_2^2 \big] \dd t,
\end{align*}
Let $\hat{u}_\theta$ denote the trained velocity estimator, obtained from i.i.d. samples $\{x_i\}_{i=1}^n$ drawn from the target data distribution $p_1$.
The following lemma establishes the convergence rate of the expected empirical risk $\mathcal{R}(\hat{u}_\theta)$ with respect to the training samples $\{x_i\}_{i=1}^n$.
\begin{lemma}[Velocity Estimation with Transformer; Theorem 4.2 of \cite{su2025high}]
\label{lem:estimation}
Let $d$ be the transformer's feature dimension.
Following standard pachify procedures \cite{peebles2023scalable}, we partition the input (data) of dimension $d_x=d \cdot L$ into a sequence of length $L$ and (transformer) feature dimension $d$.
Assume \cref{assumption:density_function_assumption_1} hold,
we have
\begin{align*}
\E_{\{x_{i}\}_{i=1}^{n}}[{\calR}(\hat{u}_{\theta})]
= O(  n^{ -\frac{1}{10d} }(\log{n})^{10d_x} ).
\end{align*}
\end{lemma}

\cref{lem:estimation} characterize how the evaluation error decrease when data sample size $n$ increases.
Next, we analyze the distribution error convergence rate of the flow matching transformers $\hat{u}_\theta$ under the TV distance. 
The following theorem quantifies how the TV distance between the true distribution and the estimated distribution decreases with increasing sample size $n$.

\begin{theorem}
[Convergence Rate under Total Variation Distance]
\label{thm:dist_esti_tv}
Let $p_1$ be the data distribution and $q_1$ be the flow matching estimated distribution. 
Let $d$ be the feature dimension.
Assume \cref{assumption:density_function_assumption_1} holds.
Then we have
\begin{align*}
    \E_{\{x_i\}_{i=1}^n}[ \text{TV}(p_1,q_1) ]
    =
   O(n^{ -\frac{1}{20d} }(\log{n})^{5d_x}).
\end{align*}
\end{theorem}

\begin{proof}
    Please see \cref{sec:proof_dist_esti_tv} for a detailed proof.
\end{proof}

We now relate our rate to the classical minimax lower bound for density estimation under Hölder smoothness.
Recall the minimax lower bound under the strong Hölder smoothness assumption:
\begin{lemma}
[Theorem 4 of \cite{yang1999information}]
\label{lem:minimax_tv}
Consider the task of estimating a probability distribution $P(x)$ with density function belonging to the space
\begin{align*}
\calP \coloneqq \big\{ p(x) \mid p(x)=f(x) \exp(-C_2\norm{x}_2^2/2): f(x) \in \calH^{\beta}(\R^{d_{x}},B), C_1 \geq f(x)\geq C \big \},
\end{align*}
Then, given i.i.d training data set 
$\{ x_i \}_{i=1}^n$ drawn from distribution $P$, 
we have
\begin{align*}
    \inf_{ \hat{P} } \sup_{ p(x) \in \calP } 
    \E_{ \{x_i\}_{i=1}^n } [\text{TV} (\hat{P}, P)] \gtrsim 
    \Omega(n^{-\frac{\beta}{d_x + 2\beta}}),
\end{align*}
where $\hat{ P }$ runs over all possible estimators constructed from the data.
\end{lemma}
We show flow matching transformers match the minimax optimal rate under specific conditions.
\begin{theorem}
[Nearly Minimax Optimality of Flow Matching Transformers]
\label{thm:minimax}
Let $C$, $C_1$ and $C_2$ be positive constants.
Assume the data distribution satisfies $p_1(x)=\exp(-C_2\norm{x}_2^2/2) \cdot f(x)$,
where $f$ belongs to H\"{o}lder space $f(x) \in \calH^{\beta}(\R^{d_{x}},B)$ (\cref{def:holder_norm_space}) and satisfies 
$ C_1 \geq f(x)\geq C$ for all $x$.
Then, within the Hölder distribution class under the Total Variation metric, the Flow Matching Transformer achieves the minimax-optimal convergence rate when $ 20d \beta =  d_x + 2\beta$.
\end{theorem}
\begin{proof}
    Please see \cref{sec:proof_minimax} for a detailed proof.
\end{proof}

\cref{lem:minimax_tv} characterizes the fundamental statistical limit of estimating smooth densities within the Hölder class.
Building on this, \cref{thm:minimax} demonstrates that the distributional convergence rate of Flow Matching Transformers scales only logarithmically worse than the classical minimax lower bound \cite{yang1999information}.
Moreover, this convergence rate matches that of diffusion models shown by \citet{fu2024unveil} using MLP networks and by \citet{hu2024statistical} using Transformer networks.
These results indicate that Flow Matching is as efficient as diffusion models in terms of minimax convergence in Hölder class under the Total Variation (TV) distance.
Consequently, our analysis provides a theoretical justification for the empirical observation that deterministic flow-based generative models achieve comparable performance to diffusion-based methods.

%% file: 3exp.tex
Our goal is to validate  \cref{lem:kl_identity} and \cref{thm:kl_bound} with toy examples.

\subsection{Validating KL Evolution Identity (\texorpdfstring{\cref{lem:kl_identity})}{}}
\label{sec:exp1}

\begin{figure}[h!]
    \centering
    \begin{minipage}[t]{0.48\linewidth}
        \centering
        \includegraphics[width=\linewidth]{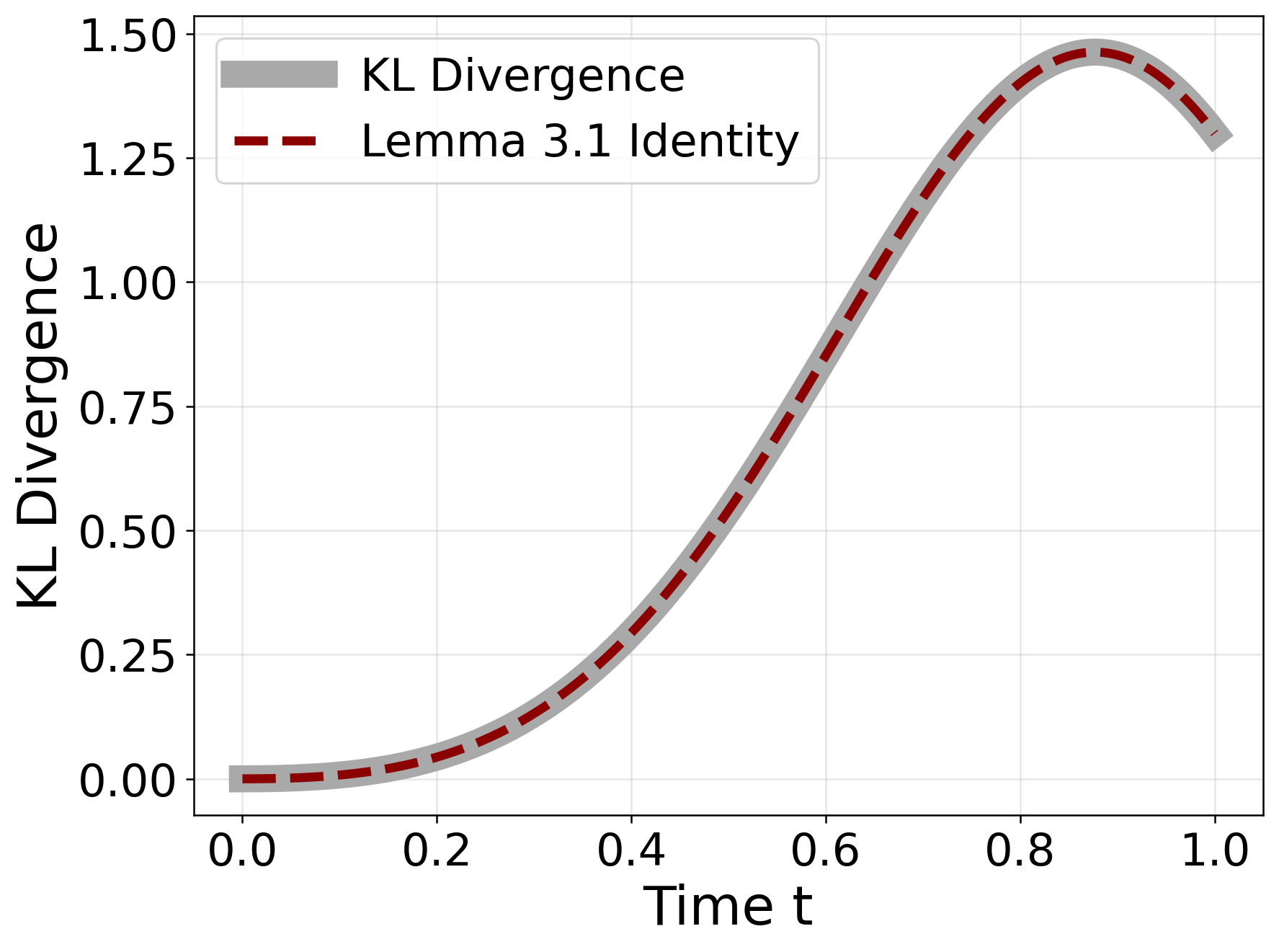}
        \vspace{-2em}
        \caption{\small\textbf{Closed-Form KL Identity (\cref{lem:kl_identity}) Verification without Learning.} Here $p_t$ evolves under $a_1(t)=\sin(\pi t)$ while $q_t$ evolves under $a_3(t)=t-\tfrac{1}{2}$.}
        \label{fig:nolearning-a1-a3}
    \end{minipage}
    \hfill
    \begin{minipage}[t]{0.48\linewidth}
        \centering
        \includegraphics[width=\linewidth]{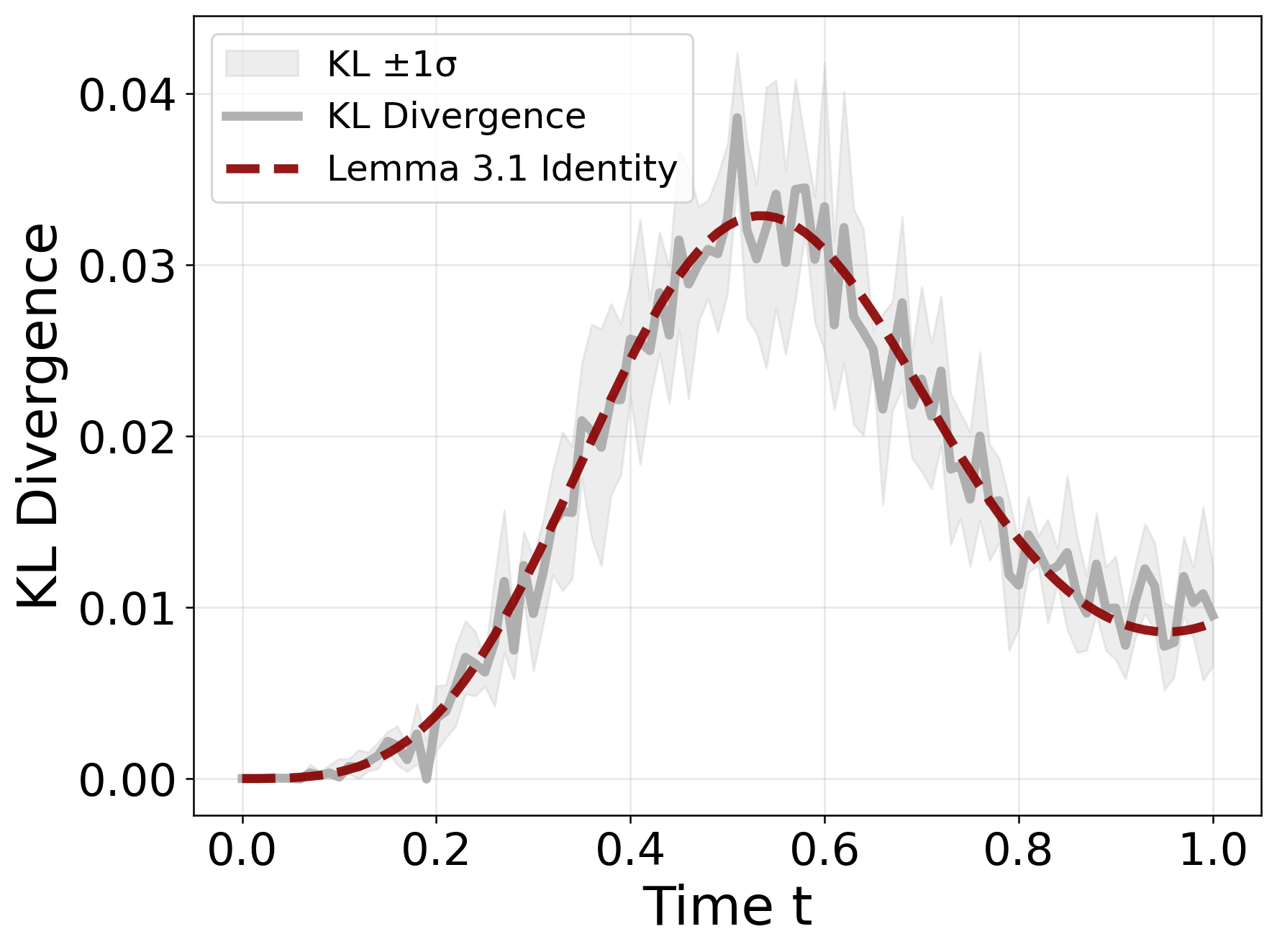}
        \vspace{-2em}
        \caption{\small\textbf{KL Identity (\cref{lem:kl_identity}) Verification with Learned Velocity Field $c_\theta$.} The model is trained on $a_2(t)=0.3\sin(2\pi t)+0.2$ until validation MSE $\le 0.05$. Sampling from $p_t$ (also under $a_2$), we compare the empirical KL divergence (dark grey) with the integrated RHS (dark red, dashed).}
        \label{fig:selflearning-a2-mse-0-05}
    \end{minipage}
\end{figure}

\paragraph{Setup.}
We validate the KL evolution identity that relates the time derivative of the divergence between two probability paths to an inner product between velocity and score mismatches:
\begin{align}
\label{eqn:exp1}
\text{KL}(p_t||q_t)
&= \int_0^t \E_{x\sim p_s}\big[\big(u(x,s)-v(x,s)\big)^\top\big(\nabla\log p_s(x)-\nabla\log q_s(x)\big)\big]  \dd s.
\end{align}
To show the generality of \cref{lem:kl_identity}, we examine two settings:
\begin{itemize}
    \item \textbf{Learn-less Flow.} \cref{lem:kl_identity} with velocity fields $v(x,s)$ given.

    \item \textbf{Learned Flow.}
    \cref{lem:kl_identity} with  velocity fields $v(x,s)$ learned from synthetic data.
\end{itemize}

\paragraph{Data.}
We work in $\mathbb{R}^2$ with $p_0=N(0,I_2)$ and a linear target field $u(x,t)=a(t)x$.
This induces $p_t=N\big(0,\sigma_p(t)^2 I_2\big)$ with $\sigma_p(t)=\exp\big(\int_0^t a(s),\dd s\big)$. We use three schedules
\begin{align*}
a_1(t) &= \sin(\pi t),\\
a_2(t) &= 0.3\sin(2\pi t)+0.2,\\
a_3(t) &= t-\tfrac{1}{2}.
\end{align*}
We evaluate on a uniform grid $t_k\in[0,1]$.

\paragraph{Model.}
We train a small MLP $v_\theta(x,t)$ by flow matching on pairs $(t,x)$ with $t\sim U[0,1]$ and $x\sim p_t$. For no learning baselines we also set $q_t$ by an analytic linear field $v(x,t)=\tilde a(t)x$.

\paragraph{Baseline (Ground Truth).}
For the left side of \eqref{eqn:exp1}, at each $t$, we draw $x^{(i)}\sim p_t$ via $x^{(i)}=\sigma_p(t)z^{(i)}$ with $z^{(i)}\sim N(0,I_2)$, compute $\log p_t(x^{(i)})$ in closed form, and obtain $\log q_t(x^{(i)})$ by integrating the backward initial-value problem $\dot x_s=-v_\theta(x_s,s)$ from $(x,t)$ to $s=0$ while accumulating $\ell(t)=\int_0^t \nabla\cdot v_\theta(x_s,s) \dd s$, then set $\log q_t(x)=\log p_0(x_0)+\ell(t)$. 
Then, the Monte Carlo estimate is
\begin{align*}
    \widehat{\text{KL}}(p_t||q_t)=\frac{1}{N}\sum_{i=1}^N \log p_t(x^{(i)})-\log q_t(x^{(i)}).
\end{align*}
For the right side integrand of \eqref{eqn:exp1}, we reuse the same samples and compute $u(x,t)=a(t)x$,  $v_\theta(x,t)$, $s_p(x,t)=\nabla\log p_t(x)=-x/\sigma_p(t)^2$, and $s_q(x,t)=\nabla_x\log q_t(x)$ by a single autograd gradient of the scalar $\log q_t(x)$ with respect to the terminal $x$. 
Then, we have the integrand estimator
\begin{align*}
\hat g(t)=\frac{1}{N}\sum_{i=1}^N \big(u(x^{(i)},t)-v_\theta(x^{(i)},t)\big)^\top\big(s_p(x^{(i)},t)-s_q(x^{(i)},t)\big).
\end{align*}
Finally, we approximate $\int_0^t \hat g(s) \dd s$ on the time grid by the trapezoidal rule.

\paragraph{Results.}
We present our results in \cref{fig:nolearning-a1-a3,fig:selflearning-a2-mse-0-05}.
\begin{itemize}
    \item \textbf{Learnless Flow (\cref{fig:nolearning-a1-a3}).}
    With $p_t$ under $a_1$ and $q_t$ under $a_3$, both sides admit closed forms and the curves coincide within numerical precision, confirming the identity.

    \item \textbf{Learned Flow (\cref{fig:selflearning-a2-mse-0-05} for $a_2$).}
    For each schedule $a_1,a_2,a_3$, we train $v_\theta$ to several validation error levels and compare the empirical left side $t\mapsto\widehat{\text{KL}}(p_t||q_t)$ against the numerical right side $t\mapsto\int_0^t\hat g(s),\dd s$. The two curves track closely across $t$.
\end{itemize}

\subsection{Validating KL Error Bounds \texorpdfstring{(\cref{thm:kl_bound})}{}}

\begin{figure}[htbp]
    \centering
    \begin{minipage}[t]{0.48\linewidth}
        \centering
        \includegraphics[width=\linewidth]{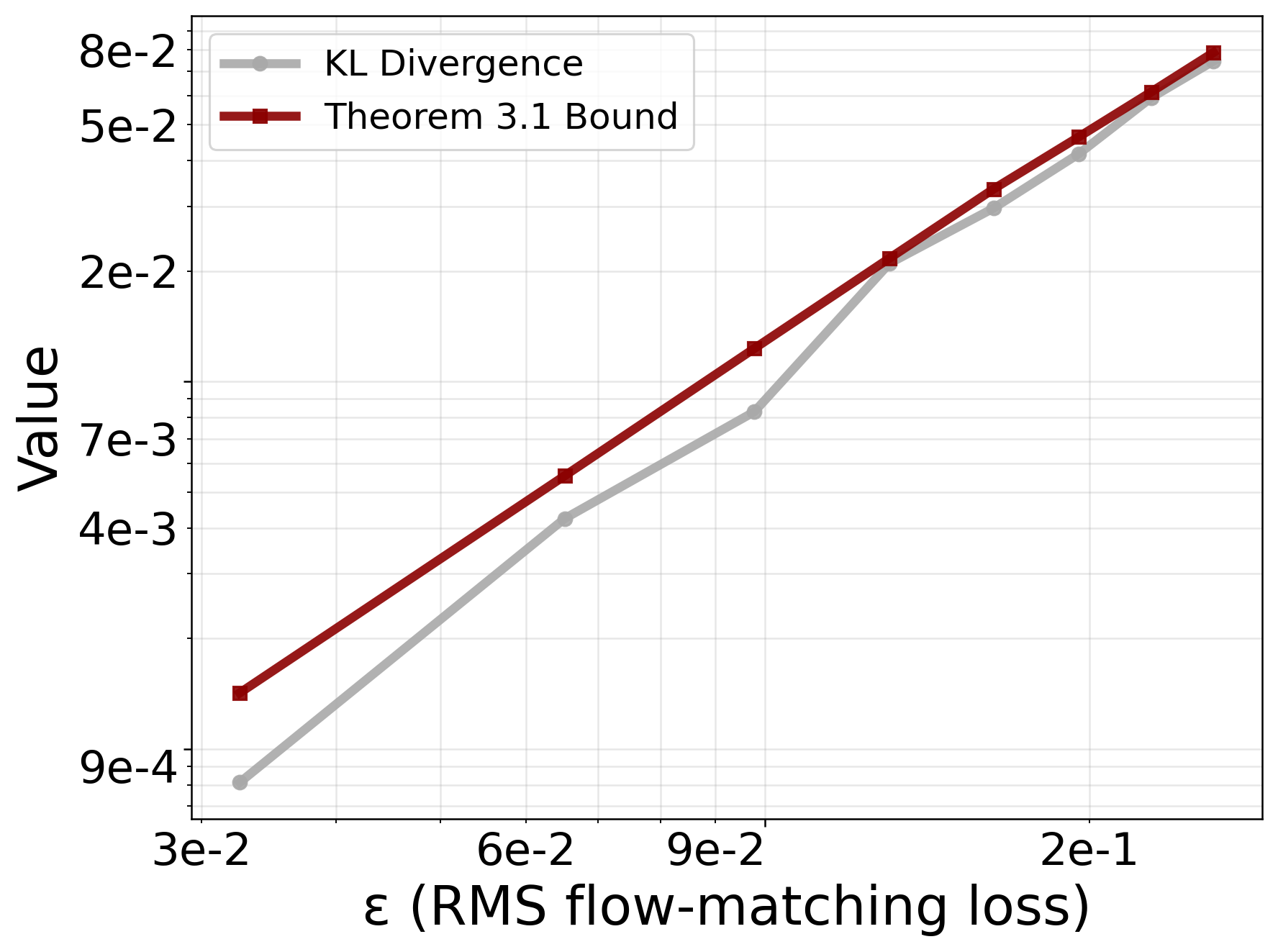}
        \vspace{-2em}
        \caption{\small\textbf{Closed-Form KL Error Bound (\cref{thm:kl_bound}) Verification.} Uses schedule $a_3$ with constant perturbations. Line plot showing $\mathrm{KL}(p_1 \Vert q_1)$ versus $\epsilon \sqrt{S}$ for synthetic velocity fields $v(x,t)=\bigl(a_3(t)+\delta(t)\bigr)x$ with $\delta(t)=\beta$, $\beta\in\{0,0.025,\dots,0.2\}$. Each point represents one perturbation configuration.}
        \label{fig:part2-nolearning-a2}
    \end{minipage}
    \hfill
    \begin{minipage}[t]{0.48\linewidth}
        \centering
        \includegraphics[width=\linewidth]{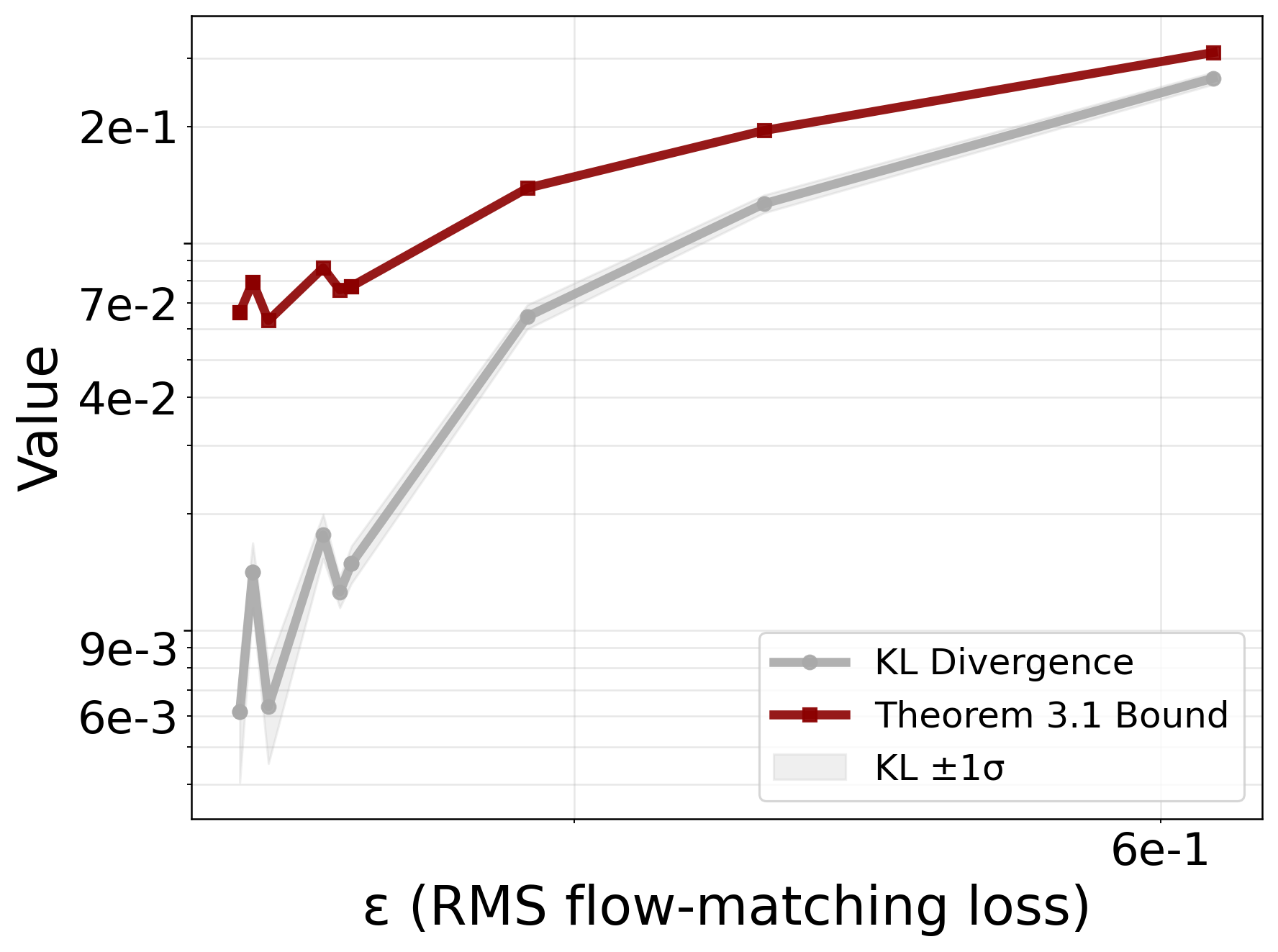}
        \vspace{-2em}
        \caption{\small\textbf{KL Error Bound (\cref{thm:kl_bound}) Verification with Learned Velocity Field.} Uses schedule $a_1$. Line plot showing $\mathrm{KL}(p_1 \Vert q_1^\theta)$ (dark grey) and $\epsilon_\theta \sqrt{S_\theta}$ (dark red) versus $\epsilon_\theta$ (RMS flow-matching loss) on log-log axes for multiple checkpoints during training. Each point represents a checkpoint at different training stages.}
        \label{fig:part2-learning-a2}
    \end{minipage}
\end{figure}

\paragraph{Setup.}
We aim to validate the learned model bound from \cref{thm:kl_bound}
\begin{align}
\label{eqn:exp2}
\text{KL}(p_1||q_1^\theta)
\le \epsilon_\theta \sqrt{\int_0^1 \mathbb{E}_{x\sim p_t}\big\lVert \nabla\log p_t(x)-\nabla\log q_t^\theta(x)\big\rVert_2^2 \dd t},
\end{align}
where $\epsilon_\theta^2=\mathbb{E}_{t\sim\mathcal{U}[0,1],x\sim p_t}\lVert v\theta(x,t)-u(x,t)\rVert_2^2$ and the time integral equals the score gap $S_\theta$.

\paragraph{Data.}
We use same Gaussian setting as \cref{sec:exp1} with $u(x,t)=a(t)x$ and schedules
\begin{align*}
a_1(t) &= \sin(\pi t),\\
a_2(t) &= 0.3\sin(2\pi t)+0.2,\\
a_3(t) &= t-\tfrac{1}{2}.
\end{align*}

\paragraph{Model.}
We train an MLP $v_\theta(x,t)$ by flow matching using Adam with cosine decay. We save a ladder of checkpoints across training to obtain a range of $\epsilon_\theta$ values.

\paragraph{Baseline (Ground Truth).}
For synthetic checks we also use $v(x,t)=(a(t)+\delta(t))x$ with constant $\delta(t)=\beta$ to produce controlled perturbations.

\paragraph{Estimators.}
For any checkpoint $\theta$:
\begin{enumerate}
    \item \textbf{LHS of \eqref{eqn:exp2}.}     
    We estimate $\text{KL}(p_1||q_1^\theta)$ by sampling $x\sim p_1$, computing $\log p_1(x)$ in closed form and $\log q_1^\theta(x)$ via a backward initial-value problem with divergence accumulation, then averaging $\log p_1(x)-\log q_1^\theta(x)$.
    \item \textbf{Flow Error $\epsilon_\theta$.} 
    We evaluate $\epsilon_\theta=\sqrt{\mathbb{E}\lVert v_\theta-u\rVert^2}$ on a uniform time grid with fresh Monte Carlo over $x\sim p_t$.
    
    \item \textbf{Score Gap $S_\theta$.} We compute $S_\theta=\int_0^1 \mathbb{E}\lVert s_p(x,t)-s_q^\theta(x,t)\rVert^2,\dd t$ by drawing common random numbers once, setting $x(t)=\sigma_p(t)z$, using $s_p(x,t)=-x/\sigma_p(t)^2$, and obtaining $s_q^\theta(x,t)$ by one autograd gradient of $\log q_t^\theta(x)$ with respect to terminal $x$.
    Then we do trapezoidal integration over $t$.
\end{enumerate}

\paragraph{Results.}
We summarize our results in \cref{fig:part2-learning-a2,fig:part2-nolearning-a2}.
\begin{itemize}

    \item \textbf{Verification with Synthetic Perturbations (\cref{fig:part2-nolearning-a2}).}
    With $v(x,t)=(a(t)+\delta(t))x$ under constant $\delta$, the same picture appears without learning. The KL divergence curve lies under the curve given by $\epsilon \sqrt{S}$, validating the bound.

    \item \textbf{Verification with Learned Velocity Fields $\epsilon$ (\cref{fig:part2-learning-a2}).}
    We plot both sides as functions of $\epsilon_\theta$ on the same axes. For each checkpoint we form the triplet $(\epsilon_\theta,\text{KL}(p_1||q_1^\theta),\epsilon_\theta\sqrt{S_\theta})$ under fixed evaluation settings and a fixed common random seed, sort by $\epsilon_\theta$, and connect the points to obtain two curves. Empirically, the left side remains below the right side across checkpoints, validating the bound.

\end{itemize}

%% file: 4conclusion.tex
We prove a deterministic, non-asymptotic KL upper bound for flow matching: if the $L_2$ flow matching loss $\mathcal{L}_{\text{FM}}$ is bounded by $\epsilon^2>0$, then
\begin{align*}
    \text{KL}(p_1||q_1) 
    \leq A_1 \epsilon + A_2 \epsilon^2,
\end{align*}
where $A_1,A_2$ depending only on regularity of the velocity fields and data scores (\cref{thm:kl_bound}). 
The proof starts from the KL evolution identity for continuity flows (\cref{lem:kl_identity}).
Our contribution is to turn this pathwise identity into an deterministic, non-asymptotic bound with explicit constants via a score-PDE analysis and Grönwall’s inequality (\cref{lem:Gronwall's}).  Leveraging the KL control, we derive Total Variation Distance convergence rates for Flow Matching Transformers (\cref{thm:dist_esti_tv}) and prove near-minimax optimality  (\cref{thm:minimax}). Experiments validate both the KL identity and the KL bound (\cref{sec:exp}). 
Overall, our results give statistical efficiency guarantees for flow matching and clarify when and how training loss controls distributional KL error.

In addition, we now prove \cref{thm:kl_bound}'s critical dependence on the assumption that the velocity fields 
$u(x,t), v(x,t)$ are differentiable and strong solutions to the continuity equations \eqref{eqn:cont_eqn_main_text}.
If we relax the $q_t$ as a non differentiable function, we construct a counterexample that for any arbitrarily small flow matching loss, the resulting terminal divergence grow without bound.

\begin{proposition}[Flow Matching Loss Fails to Control KL (Weak Version)]
\label{thm:fm_not_control_kl_weak}
Fix any constants $M>\epsilon > 0$. 
There exist probability paths $p_t(x)$ and $q_t(x)$ with $p_0(x) = q_0(x)$, and corresponding velocity fields $u(x, t)$ and $v(x, t)$ such that
for almost every $t \in [0,1]$, \cref{ass:fm_loss_bound,ass:regu_velocity} hold, and paths satisfy the weak form of continuity equations 
\begin{align*}
    \pdv{p_t(x)}{t} + \div (p_t(x) u(x, t)) = 0, \quad
    \pdv{q_t(x)}{t} + \div (q_t(x) v(x, t)) = 0.
\end{align*}
Moreover,
\begin{align*}
    \int_0^1 \E_{x \sim p_t}[\|u(x, t) - v(x, t)\|_2^2] \dd t \leq \epsilon, \quad
    \text{but} \quad
    \text{KL}(p_1 \| q_1) \geq M.
\end{align*}
\end{proposition}
\begin{proof}
    Please see \cref{sec:proof_conterex} for a detailed proof.
\end{proof}

%% file: impact.tex
By the theoretical nature of this work, we do not anticipate any negative social impact.

%% file: x_acknowledgments.tex
The authors would like to thank Minshuo Chen, Mimi Gallagher, Sara Sanchez, T.Y. Ball,  Dino Feng and Andrew Chen for valuable conversations; and Mingcheng Lu, Yi-Chen Lee, Shang Wu and Zhao Song for collaborations on related topics.

This work is in remembrance of Prof. Bi-Ming Tsai (February 9, 1965 - October 21, 2025).
Her wisdom in health, cooking, and the art of thoughtful living continues to inspire JH.

JH is partially supported by the Walter P. Murphy Fellowship and the Terminal Year Fellowship (Paul K. Richter Memorial Award) of Northwestern University.
Han Liu is partially supported by NIH R01LM1372201, NSF
AST-2421845, Simons Foundation
MPS-AI-00010513, AbbVie , Dolby and Chan Zuckerberg Biohub Chicago Spoke Award.
This research was supported in part through the computational resources and staff contributions provided for the Quest high performance computing facility at Northwestern University which is jointly supported by the Office of the Provost, the Office for Research, and Northwestern University Information Technology.
The content is solely the responsibility of the authors and does not necessarily represent the official
views of the funding agencies.

%% file: appendix.tex
{
\setlength{\parskip}{-0em}
\startcontents[sections]
\printcontents[sections]{ }{1}{}
}

\section{Related Work}
\label{sec:related_work}

\textbf{Error Bounds for Flow Matching.}
Several prior works have analyzed error bounds for fully-deterministic flow matching. \citet{albergo2022building} establish a 2-Wasserstein bound on the discrepancy between the flow-matching estimated distribution and the true data distribution with uniform Lipschitz constants. \citet{li2023towards} derive a Total Variation (TV) bound but require the estimated Stein score error to be small. \citet{benton2023error} provide a 2-Wasserstein error bound while relaxing the uniform Lipschitz condition to time-dependent regularity. However, these studies focus primarily on Wasserstein Distance or TV Distance, or rely on strong smoothness assumptions on the estimated score function.
In contrast, our work derives a non-asymptotic KL bound for flow matching that depends only on the regularity and smoothness of the data path and velocity fields, without requiring regularity of the estimated score path.

\textbf{Statistical Rates and Minimax Optimality of Flow Models.}
While error bounds study deterministic approximation for a fixed model without learning effects, statistical rate studies quantifies how error decreases with sample size.
\citet{benton2023error,albergo2022building} measure the convergence of flow models through the $L_2$ risk of the velocity field but do not provide explicit convergence rates. \citet{jiao2024convergence} derive explicit rates in the latent space of an autoencoder. However, their analysis does not account for the smoothness of the target density class. \citet{su2025theoretical} establish statistical rates for discrete flow matching by deriving a model-agnostic intrinsic error bound. \citet{fukumizu2024flow} show that flow matching achieves nearly minimax-optimal distribution estimation rates in Besov function spaces under the 2-Wasserstein distance using ReLU network architectures.
In the same vein, \citet{kunkel2025minimax} obtain similar results under the 1-Wasserstein distance using kernel density estimators. \citet{su2025high} further establish convergence rates for high-order flow matching under the 2-Wasserstein distance. However, all of these analyses are established under Wasserstein-type metrics. In contrast, we provide convergence rates for flow matching under the Total Variation (TV) distance and prove that flow matching achieves nearly minimax-optimal efficiency in TV distance, which is comparable to diffusion models.

\section{Proofs in the Main Text}
\label{sec:proofs}
\subsection{Proof of \texorpdfstring{ \cref{lem:kl_identity}}{Theorem \ref{lem:kl_identity}}}
\label{sec:proof_kl_identity}

\begin{lemma}[\cref{lem:kl_identity} Restated: KL Evolution Identity for Continuity Flows]
\label{lem:kl_identity_re}
    Let two velocities field $u(x,t),v(x,t) \in C([0,1];(C^1(\R^d))^d)$.
    Let $p_t$ and $q_t$ be two paths of differentiable probability densities on $\R^d$ evolving under the continuity equations 
    \begin{align}
    \label{eq:continuity_equations}
        \pdv{p_t(x)}{t} + \div (p_t(x) u(x, t)) = 0, \quad
        \pdv{q_t(x)}{t} + \div (q_t(x) v(x, t)) = 0,
    \end{align}
    with same initial distribution $p_0 = q_0$. 
    Then for all $t \in [0,1]$,
    \begin{align}
    \label{eq:kl_transport_id}
    \dv{}{t} \text{KL}(p_t||q_t)
    = \E_{x \sim p_t}[ \left(u(x,t)-v(x,t) \right)^\top 
    \left( \nabla \log p_t(x) - \nabla \log q_t(x) \right)].    
    \end{align}
\end{lemma}

\begin{proof}
Let all probability density functions and velocity firlds defined on $\Omega = \R^d$. 
Then for any differentiable function $f_t(x): \R^d \rightarrow \R$, we have
\begin{align}
    & ~ \dv[]{}{t} \int_\Omega p_t(x) f_t(x) \dd x \nonumber\\
    = & ~ \int_\Omega \pdv{p_t(x)}{t} f_t(x) \dd x + \int \pdv{f_t(x)}{t} p_t(x) \dd x \annot{By the product rule} \\
    = & ~ -\int_\Omega \div 
    (p_t(x) u(x,t)) f_t(x) \dd x + \int_\Omega \pdv{f_t(x)}{t} p_t(x) \dd x \annot{By replacing the first term by continuity equation \eqref{eq:continuity_equations}}\\
    = & ~ \int_\Omega (p_t(x) u(x,t))^\top \nabla f_t(x) \dd x - \int_{\partial \Omega} p_t(x) u(x,t) f_t(x) \cdot  (\hat{n} \dd S) + \int_\Omega \pdv{f_t(x)}{t} p_t(x) \dd x \annot{By divergence theorem}\\
    = & ~ \int_\Omega (p_t(x) u(x,t))^\top \nabla f_t(x) \dd x + \int_\Omega \pdv{f_t(x)}{t} p_t(x) \dd x \notag\\
    = & ~ \int_\Omega p_t(x) [\pdv{}{t} f_t(x) + u(x,t)^\top \nabla f_t(x)] \dd x.
    \label{eq:derivative_f}
\end{align}

Now we let $f_t(x) = \log \frac{p_t(x)}{q_t(x)}$, then the derivative of KL divergence between $p_t$ and $q_t$ becomes
\begin{align}
    \dv{}{t} \text{KL}(p_t||q_t)
    = & ~ \dv{}{t} \int_\Omega p_t(x) f_t(x) \dd x \annot{By the definition of KL divergence}\\ 
    = & ~ \int_\Omega p_t(x) [\underbrace{\pdv{}{t} \log \frac{p_t(x)}{q_t(x)}}_{:=(I)} + \underbrace{u(x,t)^\top \nabla \log \frac{p_t(x)}{q_t(x)}}_{:= (II)}] \dd x. 
    \label{eq:kl_integral_1}
\end{align}

Then we calculate (I) and (II) terms in \eqref{eq:kl_integral_1}.
From the continuity equations \eqref{eq:continuity_equations}, we have
\begin{align*}
    \pdv{\log p_t(x)}{t}
    = & ~ -u(x,t)^\top \nabla \log p_t(x) - \div u(x,t), \\
    \pdv{\log q_t(x)}{t}
    = & ~ -v(x,t)^\top \nabla \log q_t(x) - \div v(x,t).
\end{align*}
Consequently, (I) becomes
\begin{align}
\label{eq:first_term_in_int}
    \pdv{}{t} \log \frac{p_t(x)}{q_t(x)}
    = - \div u(x,t) + \div v(x,t) - u(x,t)^\top \nabla \log p_t(x) + v(x,t)^\top \nabla \log q_t(x).
\end{align}
For (II), by the property of logarithmic function, we have
\begin{align}
\label{eq:second_term_in_int}
    u(x,t)^\top \nabla \log \frac{p_t(x)}{q_t(x)}
    = u(x,t)^\top [\nabla \log p_t(x) - \nabla \log q_t(x)].
\end{align}

Substituting \eqref{eq:first_term_in_int} and \eqref{eq:second_term_in_int} into \eqref{eq:kl_integral_1}, the derivative of KL divergence becomes
\begin{align*}
    \dv{}{t} \text{KL}(p_t||q_t)
    = & ~ \int_\Omega p_t(x) [\pdv{}{t} \log \frac{p_t(x)}{q_t(x)} + u(x,t)^\top \nabla \log \frac{p_t(x)}{q_t(x)}] \dd x \\
    = & ~ \int_\Omega p_t(x) [- \div u(x,t) + \div v(x,t) + (-u(x,t)+v(x,t))^\top \nabla \log q_t(x)] \dd x.
\end{align*}

Notice that, for any $C^1$ vector field $a(x): \R^d \rightarrow \R^d$ and the probability density function $p_t(x)$,
\begin{align}
    \E_{x \sim p_t}[\div a(x)]
    = & ~\int_{\Omega} (\div a(x)) p_t(x) \dd x \annot{By the definition of expectation}\\
    = & ~ \int_{\partial \Omega}  a(x) p_t(x) \cdot (\hat{n}
    \dd S) -\int_{\Omega} a(x)^\top \nabla p_t(x) \dd x \annot{By divergence theorem} \\
     = & ~ - \int_{\Omega} a(x)^\top \nabla p_t(x) \dd x \notag\\
     = & ~ - \int_{\Omega} a(x)^\top \nabla \log p_t(x) p_t(x) \dd x  \annot{By $\nabla \log p_t(x) = \nabla p_t(x) / p_t(x)$} \\
     = & ~ - \E_{x \sim p_t}[a(x)^\top \nabla \log p_t(x)].
     \label{eq:expect_div_trick}
\end{align}

Let $a(x) = - u(x,t) + v(x,t)$.
Then the derivative of KL divergence becomes
\begin{align*}
    \dv{}{t} \text{KL}(p_t||q_t)
    = & ~ \int_\Omega p_t(x) [- \div u(x,t) + \div v(x,t) + (-u(x,t)+v(x,t))^\top \nabla \log q_t(x)] \dd x \\
    = & ~ \int_\Omega p_t(x) [(u(x,t)-v(x,t))^\top (\nabla \log p_t(x) - \nabla \log q_t(x)] 
    \dd x \annot{By \eqref{eq:expect_div_trick}}\\
    = & ~ \E_{x \sim p_t}[(u(x,t)-v(x,t))^\top (\nabla \log p_t(x) - \nabla \log q_t(x))].
\end{align*}
This completes the proof.
\end{proof}

\subsection{Proof of \texorpdfstring{ \cref{thm:kl_bound}}{Theorem \ref{thm:kl_bound}}}
\label{sec:proof_kl_bound}
\begin{theorem}[\cref{thm:kl_bound} Restated: KL Error Bounds for Flow Matching]
\label{thm:kl_bound_re}
Assume \cref{ass:fm_loss_bound,ass:regu_score,ass:regu_velocity} hold. 
If the initial distributions $p_0=q_0$, then 
\begin{align*}
    \text{KL}(p_1||q_1)
    \leq & ~ \epsilon ~ 
    \sqrt{\int_0^1 \E_{x \sim p_t}
    [\|\nabla \log p_t(x) - \nabla \log q_t(x)\|_2^2] \dd t}. 
\end{align*}

Let the lipschitz constants $U_p(t),B_p(t),K(t),L(t),M(t),H(t)$ be as defined in \cref{ass:regu_score} and \cref{ass:regu_velocity}. 
Then, we bound the terminal KL divergence by 
\begin{align*}
    \text{KL}(p_1||q_1)
    \leq A_1 \epsilon + A_2 \epsilon^2,
\end{align*} 
where
\begin{align*}
    A_1:= & ~ \exp{\int_0^1 L(t) + K(t) + B_p(t) M(t)\dd t} \cdot \int_0^1 2L(t)B_p(t) + 2H(t)\dd t, \\
    A_2:= & ~ \exp{\int_0^1 L(t) + K(t) + B_p(t) M(t)\dd t} \cdot \sqrt{\int_0^1 U_p(t)^2 \dd t}.
\end{align*}
\end{theorem} 
\begin{proof}
    Assuming the second moment of stein score error is bounded (\cref{ass:regu_score}), we bound the KL divergence  $\text{KL}(p_t||q_t)$ with the Cauchy-Schwarz Inequality for expectations:
\begin{align*}
    \text{KL}(p_1||q_1)
    = & ~ \int_0^1 \E_{x \sim p_t}[(u(x,t)-v(x,t))^\top (\nabla \log p_t(x) - \nabla \log q_t(x)] \dd t \\
    \leq & ~ \int_0^1 \sqrt{\E_{x \sim p_t}[\|u(x,t)-v(x,t)\|^2_2]} \cdot \sqrt{\E_{x \sim p_t}[\|\nabla \log p_t(x) - \nabla \log q_t(x)\|^2_2]}\dd t,
\end{align*}
where the first line is by the KL difference for continuity flows \eqref{eq:kl_ter_id}, and the second line is by Cauchy-Schwarz Inequality.
Then, by Cauchy-Schwarz Inequality on $L_2$ space, we have
\begin{align}
\label{eq:kl_bound_score}
    \text{KL}(p_1||q_1)
    \leq & ~ \sqrt{\int_0^1 \E_{x \sim p_t}[\|u(x,t)-v(x,t)\|^2_2] \dd t} \cdot \sqrt{\int_0^1 \E_{x \sim p_t}[\|\nabla \log p_t(x) - \nabla \log q_t(x)\|^2_2] \dd t} \notag \\
    \leq & ~ \epsilon ~ 
    \sqrt{\int_0^1 \E_{x \sim p_t}
    [\|\nabla \log p_t(x) - \nabla \log q_t(x)\|^2_2] \dd t},
\end{align}
where the last step follows from \cref{ass:fm_loss_bound}.

To bound the mean-squared expected score discrepancy term $\E_{x \sim p_t}[\|\nabla \log p_t(x) - \nabla \log q_t(x)\|^2_2]$, we exploit the score PDE induced by \eqref{eq:derivative_f} and the continuity equations \eqref{eq:continuity_equations}:
\begin{align*}
    \pdv{\nabla \log p_t(x)}{t}
    = & ~ \nabla[-u(x,t)^\top \nabla \log p_t(x) - \div u(x,t)]
    \\
    = & ~ -(\nabla u(x,t))^\top \nabla \log p_t(x)
    - \nabla (\nabla \log p_t(x)) u(x,t)
    - \nabla (\div u(x,t)),
\end{align*}
where the first line is by the continuity equations \eqref{eq:continuity_equations}.

For simplicity, we define the  scores for true probability paths $p_t$ as and estimated probability paths $q_t$ as $s_p := \nabla \log p_t(x)$ and $s_q := \nabla \log q_t(x)$, respectively.

Then, by symmetry and above equation, we have
\begin{align}
\label{eq:score_pde_1}
    \pdv{s_p}{t} = -(\nabla u(x,t))^\top s_p - \nabla s_p u(x,t) - \nabla (\div u(x,t)), \\
\label{eq:score_pde_2}
    \pdv{s_q}{t} = -(\nabla v(x,t))^\top s_q - \nabla s_q v(x,t) - \nabla (\div v(x,t)).
\end{align}

For simplicity, let $\Delta s := s_p - s_q$ denote the stein score error.
Then by \eqref{eq:score_pde_1} and \eqref{eq:score_pde_2}, we have
\begin{align}
\label{eq:score_error_pde}
    \pdv{\Delta s}{t} = - (\nabla u)^\top \Delta s - (\nabla u -\nabla v )^\top s_q -\nabla(\Delta s) u - \nabla s_q (u-v)- \nabla (\div u - \div v).
\end{align}
By $s_q = s_p - \Delta s$,  \eqref{eq:score_error_pde} becomes
\begin{align}
    & ~ \pdv{\Delta s}{t} \nonumber\\
    = & ~- (\nabla u)^\top \Delta s - (\nabla u -\nabla v )^\top ( s_p - \Delta s) -\nabla(\Delta s) u - (\nabla s_p - \nabla \Delta s)(u-v)- \nabla (\div u - \div v) \nonumber\\
    = & ~ -(\nabla v)^\top \Delta s - (\nabla u -\nabla v )^\top s_p -\nabla(\Delta s) u - (\nabla s_p - \nabla \Delta s)(u-v)- \nabla (\div u - \div v) .
    \label{eq:score_pde_expand}
\end{align}

Now differentiate the expected mean square Stein score error and get
\begin{align}
\label{eqn:dtds_2}
    & ~ \dv{}{t} 
    \E_{x \sim p_t}
    [\frac{1}{2} \|\Delta s\|_2^2]\\
    = & ~ \dv{}{t} \int_\Omega p_t(x) \cdot \frac{1}{2} \|\Delta s\|_2^2 \dd x \annot{By definition of expectation}\\
    = & ~ \int_\Omega p_t(x) [\pdv{}{t} \frac{1}{2} \|\Delta s\|_2^2 + u(x,t)^\top \nabla \frac{1}{2} \|\Delta s\|_2^2] \dd x \annot{By replacing $f_t(x)$ by $\frac{1}{2} \|\Delta s\|_2^2$ in \eqref{eq:derivative_f}}\\
    = & ~ \int_\Omega p_t(x) [(\pdv{\Delta s}{t} + \nabla(\Delta s) u)^\top \Delta s] \dd x \annot{By matrix calculus}\\
    = & ~ \underbrace{- \E_{x \sim p_t}[\Delta s^\top (\nabla v) \Delta s]}_{:= (I)}
    \underbrace{- \E_{x \sim p_t}[(s_p^\top (\nabla u - \nabla v) \Delta s]}_{:=(II)}
    \underbrace{- \E_{x \sim p_t}[(u-v)^\top (\nabla s_p) \Delta s]}_{:= (III)} \\
    & ~ 
     \underbrace{- \E_{x \sim p_t}[(u-v)^\top (- \nabla \Delta s) \Delta s]}_{:= (IV)} 
    \underbrace{- \E_{x \sim p_t}[\nabla (\div u - \div v)^\top 
    \Delta s]}_{:= (V)}.
    \annot{By substituting the score PDE \eqref{eq:score_pde_expand}}
\end{align}
We bound each terms with Lipchitz constants separately.
\begin{itemize}
    \item For (I), we have
    \begin{align*}
            - \E_{x \sim p_t}[\Delta s^\top (\nabla u) \Delta s] 
        \leq  & ~ L(t)
        \E_{x \sim p_t}[\|\Delta s\|_2^2].
    \end{align*}
    \item For (II), we have 
    \begin{align*}
        - \E_{x \sim p_t}[(s_p^\top (\nabla u - \nabla v) \Delta s]
        \leq  & ~ 2L(t)
        \E_{x \sim p_t}[ \|s_p\|_2 \cdot \|\Delta s\|_2] \\
        \leq & ~ 2L(t)  B_p(t) \sqrt{\E_{x \sim p_t}[\|\Delta s\|_2^2]}. \annot{By Cauchy-Schwarz Inequality and \eqref{ass:regu_score}}
    \end{align*}
    \item For (III), we have
    \begin{align*}
        -\E_{x \sim p_t}[(u-v)^\top (\nabla s_p) \Delta s]
        \leq & ~ \E_{x \sim p_t}[\|(u-v)^\top (\nabla s_p) \Delta s\|_2]\\
        \leq & ~ U_p(t) \E_{x \sim p_t}[\|(u-v)^\top \Delta s\|_2]. \annot{By the definition of $U_p(t)$ follows \cref{ass:regu_score}} \\
        \leq & ~ U_p(t) \epsilon(t) \sqrt{\E_{x \sim p_t}[\|\Delta s\|_2^2]}.
        \annot{By Cauchy-Schwarz Inequality}
    \end{align*}
    \item For (IV), we have
    \begin{align*}
            \E_{x \sim p_t}[(u-v)^\top (\nabla \Delta s) \Delta s]
        = & ~\E_{x \sim p_t}[(u-v)^\top \nabla(\frac{1}{2} \|\Delta s\|_2^2)] \annot{By chain rule}\\
        = & ~ \int_{\R^d} p_t(x) (u-v)^\top \nabla(\frac{1}{2} \|\Delta s\|_2^2)\dd x \annot{By the definiton of expectation}\\
        = & ~ - \int_{\R^d} \frac{1}{2} \|\Delta s\|_2^2 \div (p_t(x) (u-v))\dd x \annot{By the divergence theorem} \\
        = & ~ -\int_{\R^d} p_t(x) \cdot \frac{1}{2} \|\Delta s\|_2^2 (\div(u-v) + (u-v)^\top s_p) \dd x \annot{By the product rule}\\
        = & ~ -\E_{x \sim p_t}[\frac{1}{2} \|\Delta s\|_2^2 \left((\div u - \div v) + (u-v)^\top s_p\right)] \\
        \leq & ~ K(t) \E_{x \sim p_t}[\|\Delta s\|_2^2] + \frac{1}{2} B_p(t) \E_{x \sim p_t}[\|\Delta s\|_2^2 \cdot \|u-v\|_2] \annot{By the definition of $K(t)$ and $B_p(t)$ follow \cref{ass:regu_velocity}} \\
        \leq & ~ K(t) \E_{x \sim p_t}[\|\Delta s\|_2^2] + B_p(t) M(t) \E_{x \sim p_t}[\|\Delta s\|_2^2] . \annot{By definition of $M(t)$ follows \cref{ass:regu_velocity}} 
    \end{align*}
    \item For (V), we have
    \begin{align*}
        - \E_{x \sim p_t}[\nabla (\div u - \div v)^\top 
        \Delta s] 
        \leq & ~ 2H(t) \sqrt{\E_{x \sim p_t}[\|\Delta s\|_2^2]}. \annot{By Cauchy-Schwarz Inequality}
    \end{align*}

\end{itemize}

Collecting all above into \eqref{eqn:dtds_2}, dividing both sides by $\sqrt{\E_{x \sim p_t}[\|\Delta s\|_2^2]}$, we bound the derivative of $\sqrt{\E_{x \sim p_t}[\|\Delta s\|_2^2]}$ as 
\begin{align*}
    & ~ \dv{}{t} \sqrt{\E_{x \sim p_t}[\|\Delta s\|_2^2]} \\
    \leq & ~ 
    \left(L(t) + K(t)+ B_p(t) M(t)\right)
    \sqrt{\E_{x \sim p_t}[\|\Delta s\|_2^2]} + 2L(t)B_p(t) + U_p(t)\epsilon(t) +2H(t).
\end{align*}

Applying Gr\"{o}nwall's Inequality \cref{lem:Gronwall's}, we have
\begin{align*}
    \sqrt{\E_{x \sim p_t}[\|\Delta s\|_2^2]}
    \leq & ~ \sqrt{\E_{x \sim p_0}[\|\Delta s\|_2^2]} e^{C_1(t)} + \int_0^t e^{C_1(t) - C_1(\tau)} C_2(\tau) \dd \tau, \\
    = & ~ \int_0^t e^{C_1(t) - C_1(\tau)} C_2(\tau) \dd \tau , \annot{By $p_0=q_0$}
\end{align*}
where $C_1(t) = \int_0^t L(s) + K(s) +  B_p(s) M(s)\dd s$ and $C_2(t) = 2L(t)B_p(t) + U_p(t)\epsilon(t) + 2H(t)$.

Square both sides and take integral over $t \in [0,1]$,
\begin{align*}
    \int_0^1 \E_{x \sim p_t}[\|\Delta s\|_2^2] \dd t
    \leq & ~ \int_0^1 \left(\int_0^t e^{C_1(t) - C_1(\tau)} C_2(\tau)  \dd \tau \right)^2 \dd t \\
    \leq & ~ \int_0^1 \left(\int_0^t e^{C_1(1)} C_2(\tau)  \dd \tau \right)^2 \dd t \annot{By $C_1(t) - C_1(\tau)\leq C_1(1)$}\\
    \leq & ~ \left(\int_0^1 e^{C_1(1)} C_2(\tau)  \dd \tau \right)^2 \annot{By $ C_2(t) \geq 0$}.
\end{align*}
Consequently, we bound the KL divergence in terms of Lipschitz Constants as
\begin{align*}
    \text{KL}(p_1||q_1)
    \leq & ~ \epsilon ~ 
    \sqrt{\int_0^1 \E_{x \sim p_t}
    [\|\nabla \log p_t(x) - \nabla \log q_t(x)\|_2^2] \dd t} \annot{By \eqref{eq:kl_bound_score}} \\
    \leq & ~ \epsilon \left(\int_0^1 e^{C_1(1)} C_2(\tau)  \dd \tau \right)\\
    \leq & ~ \epsilon \cdot e^{C_1(1)} \left(\int_0^1  2L(t)B_p(t) + U_p(t)\epsilon(t) + 2H(t)  \dd t \right) \annot{By the definition of $C_2(t)$} \\
    \leq & ~ \epsilon \cdot e^{C_1(1)} \left(\int_0^1  \left(2L(t)B_p(t) + 2H(t)\right) \dd t + \sqrt{\int_0^1 U_p(t)^2 \dd t} \epsilon \right)  \annot{By the definition of $\epsilon_t$ follows \cref{ass:fm_loss_bound} and Cauchy-Schwarz Inequality} \\
    \leq & ~ A_1\epsilon + A_2 \epsilon^2,
\end{align*}
where the last step follows from
defining 
\begin{align*}
   A_1 := & ~ e^{C_1(1)} \int_0^1  \left(2L(t)B_p(t) + 2H(t)\right) \dd t \\
   A_2 := & ~ e^{C_1(1)} \sqrt{\int_0^1 U_p(t)^2 \dd t}. 
\end{align*}
This completes the proof.
\end{proof}

\subsection{Proof of \texorpdfstring{ \cref{thm:fm_not_control_kl_weak}}{Theorem \ref{thm:fm_not_control_kl_weak}}}
\label{sec:proof_conterex}
\begin{theorem}[\cref{thm:fm_not_control_kl_weak} Restated: Flow Matching Loss Fails to Control KL (Weak Solution Version)]
\label{thm:fm_not_control_kl_weak_re}
Fix any constants $M>\epsilon > 0$. 
There exist probability paths $p_t(x)$ and $q_t(x)$ with $p_0(x) = q_0(x)$, and corresponding velocity fields $u(x, t)$ and $v(x, t)$ such that
for almost every $t \in [0,1]$, \cref{ass:fm_loss_bound,ass:regu_velocity} hold, and paths satisfy the continuity equations 
\begin{align*}
    \pdv{p_t(x)}{t} + \div (p_t(x) u(x, t)) = 0, \quad
    \pdv{q_t(x)}{t} + \div (q_t(x) v(x, t)) = 0.
\end{align*}
Moreover,
\begin{align*}
    \int_0^1 \E_{x \sim p_t}[\|u(x, t) - v(x, t)\|_2^2] \dd t \leq \epsilon, \quad
    \text{but} \quad
    \text{KL}(p_1 \| q_1) \geq M.
\end{align*}
\end{theorem}

\begin{proof}
Let $p_t \equiv p_0 = N(0,I)$ with $u \equiv 0$.
Then the continuity equation $\pdv{p_t}{t} + \div (p_t u_t)=0$ is satisfied.
Define the helping function $\psi_t(x) = a(t) b^\top x$, where $b \in \R^d$ is a constant vector and $a(t)$ is a scalar function satisfy following ordinary differential equation for some $\delta > 0$ to be determined:
\begin{align}
\label{eq:a_ode}
    a'(t) := \dv{}{t}a(t) = \delta a(t), \quad \text{almost every}\quad t \in [0,1].
\end{align}
Define estimated probability path $q_t$ via $p_t$ by
\begin{align*}
    q_t(x) = Z_t^{-1} p_t(x) \exp(-\psi_t(x)),
    \quad
    Z_t = \int p_t e^{-\psi_t} \dd x \in (0,\infty),
\end{align*}
which implies that
\begin{align}
\label{eq:score_q}
    \nabla \log q_t(x) = \nabla \log p_t(x) - \nabla \psi_t(x).
\end{align}

Next, we define the constant estimated velocity field $v(x,t)$ as
\begin{align}
\label{eq:v_def}
    v(x,t) :=
    -\delta \nabla \psi_t(x) = -\delta a(t)b.
\end{align}
Then we check the estimated probability paths $q_t$ and the estimated velocity $v$ satisfy the continuity equation for almost every $t \in [0,1]$.
Notice that $\log q_t = - \log Z_t + \log p_0 - \psi_t(x)$, therefore 
\begin{align}
\label{eq:partial_q}
    \pdv{}{t} q_t(x) = q_t(x)\cdot \pdv{}{t} \log q_t(x)
    = q_t(x) \cdot (- \pdv{}{t} \log Z_t - \pdv{}{t} \psi_t(x)).
\end{align}
Consequently, for almost everywhere $t \in [0,1]$, the continuity equation becomes
\begin{align*}
    & ~ \pdv{}{t} q_t(x) + \div(q_t(x) v(x,t)) \\
    = & ~ q_t(x) \cdot \left(- \pdv{}{t} \log Z_t - \pdv{}{t} \psi_t(x)\right) + \div(q_t(x) v(x,t))  \annot{By \eqref{eq:partial_q}} \\
    = & ~ q_t(x) \cdot \left(- \pdv{}{t} \log Z_t - \pdv{}{t} \psi_t(x) + \div v(x,t) + \nabla \log q_t(x)^\top v(x,t) \right)
    \annot{By the product rule}\\
    = & ~ q_t(x) \cdot \left(- \pdv{}{t} \log Z_t - \pdv{}{t} \psi_t(x) 
    - \delta a(t) \nabla \log q_t(x)^\top b \right)
    \annot{By \eqref{eq:v_def}, \eqref{eq:score_q} and $\nabla \log p_t(x)=0$} \\
    = & ~ q_t(x) \cdot \left(- \pdv{}{t} \log Z_t - a'(t) b^\top x 
    - \delta a(t) b^\top (\nabla p_0(x) - \nabla \psi_t(x)) \right)
    \annot{By $\psi_t(x)=a(t)b^\top x$ and $\log q_t = - \log Z_t + \log p_0 - \psi_t(x)$} \\
    = & ~ q_t(x) \cdot \left(- \pdv{}{t} \log Z_t - a'(t) b^\top x 
    + \delta a(t) b^\top (x + a(t)b) \right) \annot{By $p_0 = N(0,I)$ and $\nabla \psi_t(x)=a(t)b$} \\
    = & ~ q_t(x) \cdot \left(- a'(t) b^\top x 
    + \delta a(t) b^\top x + \delta a(t)^2 \|b\|^2_2 - \pdv{}{t} \log Z_t \right).
\end{align*}

Following the definition of $a(t)$ \eqref{eq:a_ode}, $ - a'(t) b^\top x 
+ \delta a(t) b^\top x = 0$. 
For another term, recall that $Z_t = \int_{\R^d} p_t(x) e^{-\psi_t(x)} \dd x$ and $p_t = p_0 = N(0,I)$, which implies that for almost every $t \in [0,1]$,
\begin{align*}
    & ~ \delta a(t)^2 \|b\|^2_2 - \pdv{}{t} \log Z_t \\
    = & ~ \delta a(t)^2 \|b\|^2_2 - \pdv{}{t} \log  \int_{\R^d} p_0(x) e^{-\psi_t(x)} \dd x \annot{By the defintion of $Z_t$} \\
    = & ~ \delta a(t)^2 \|b\|^2_2 - \pdv{}{t} \log \E_{x \sim p_0}[e^{-\psi_t(x)}]  \\
    = & ~ \delta a(t)^2 \|b\|^2_2 - \pdv{}{t} \log \E_{x \sim p_t}[e^{-a(t) b^\top x}] \annot{By the definition of $\psi_t(x)= a(t) b^\top x$} \\
    = & ~ \delta a(t)^2 \|b\|^2_2 - \pdv{}{t}
    \frac{1}{2 }a(t)^2 \|b\|^2_2
    \annot{By $\E_{x \sim N(0,I)}[e^{t^\top x}] = e^{ 
    \frac{1}{2} t^\top t}$} \\
    = & ~ \delta a(t)^2 \|b\|^2_2 - a'(t) a(t) \|b\|^2_2 \annot{By the chain rule} \\
    = & ~ 0 \annot{By $a'(t) = \delta a(t)$ follows \eqref{eq:a_ode}}.
\end{align*}

Therefore the continuity equation $\pdv{q_t(x)}{t} + \div (q_t(x) v(x, t)) = 0$ holds almost everywhere.
Then we calculate the KL bounds between $p_t(x)$ and $q_t(x)$,
\begin{align}
    \dv{}{t} \text{KL}(p_t\|q_t) 
    = & ~ \E_{x \sim p_t}[\left(u(x,t) - v(x,t)\right) \cdot (\nabla \log p_t(x) - \nabla \log q_t(x))] \annot{By KL evolution identity \cref{lem:kl_identity}} \\
    = & ~ \E_{x \sim p_t}[v(x,t) \nabla \log q_t(x)] \annot{By $u \equiv 0$} \\
    = & ~ \E_{x \sim p_t}[\delta a(t)^2 \|b\|_2^2] \annot{By $v=-\delta a(t) b$ and $\nabla \log q_t(x) = -\nabla \psi_t(x)$ follows \eqref{eq:score_q}} \\
    = & ~ \delta a(t)^2 \|b\|_2^2.
    \label{eq:derivative_kl}
\end{align}
Following the ordinary
differential equation \eqref{eq:a_ode}, we define $a(t)$ as
\begin{align}
\label{eq:a_def}
    a(t) = 
    \begin{cases}
        0, & 0 \leq t \leq \tau, \\
        \eta e^{\delta(t-\tau)} & \tau < t \leq 1.
    \end{cases}
\end{align}
Integrating over $t \in [0,1]$ and using $p_0 = q_0$, we obtain
\begin{align*}
    \text{KL}(p_1\|q_1)
    = & ~ \int_0^1 \dv{}{t} \text{KL}(p_t\|q_t) \dd t \\
    = & ~ \delta \|b\|_2^2 \int_0^1 a(t)^2 \dd t \annot{By \eqref{eq:derivative_kl}}\\
    = & ~ \delta \|b\|_2^2 
    \int_\tau^1 \eta^2 e^{2\delta(t-\tau)} \dd t \annot{By the definition of $a(t)$ follows \eqref{eq:a_def}}\\
    = & ~ \delta \|b\|_2^2 
    \frac{\eta^2}{2\delta} (e^{2\delta(t-\tau)}-1).
\end{align*}
Define $J:= \int_0^1 a(t)^2 \dd t= \frac{\eta^2}{2\delta} (e^{2\delta(t-\tau)}-1)$, then for the flow matching loss 
\begin{align*}
    \int_0^1 \E_{x \sim p_t}[\|u(x, t) - v(x, t)\|_2^2] \dd t
    = \int_0^1 \delta^2 a(t)^2 \|b\|_2^2 \dd t 
    = \delta^2 \|b\|_2^2 J
    = \delta \cdot \text{KL}(p_1\|q_1).
\end{align*}
Given any $M>\epsilon>0$, let $\delta = {\epsilon}/{M}$,
then there exist a unique $\eta>0$ satisfy $J={\epsilon}/{\delta^2 \|b\|_2^2}$.
Then by our construction, we have flow matching loss
\begin{align*}
    \int_0^1 \E_{x \sim p_t}[\|u(x, t) - v(x, t)\|_2^2]=\delta^2 \|b\|_2^2 J=\epsilon,
\end{align*}
and $\text{KL}(p_1\|q_1) = \epsilon/\delta = M$.
This completes the proof.
\end{proof}

\begin{remark}[Interpretation]
The construction aligns a small velocity disparity $u - v$ with a large score gap $\nabla \log p_t - \nabla \log q_t = \nabla \psi_t$. 
The flow-matching loss controls only $\E_{x \sim p_t}\|u - v\|^2$, whereas the KL evolution depends on their \emph{correlation} $(u - v) \cdot (\nabla \log p_t - \nabla \log q_t)$. 
Without an independent control on the score gap (e.g., its $L_2(p_t)$ norm), the KL divergence can increase arbitrarily even when the flow-matching loss remains small.
\end{remark}

\subsection{Proof of \texorpdfstring{ \cref{thm:dist_esti_tv}}{Theorem \ref{thm:dist_esti_tv}}}
\label{sec:proof_dist_esti_tv}

\begin{theorem}
[\cref{thm:dist_esti_tv} Restated: Distribution Estimation under Total Variation Distance]
\label{thm:dist_esti_tv_re}
Let $p_1$ be the data distribution and $q_1$ be the flow matching estimated distribution. 
Let $d$ be the feature dimension.
Assume \cref{assumption:density_function_assumption_1} holds.
Then we have
\begin{align*}
    \E_{\{x_i\}_{i=1}^n}[ \text{TV}(p_1,q_1) ]
    =
    O(  n^{ -\frac{1}{20d} }(\log{n})^{5d_x} ).
\end{align*}

\end{theorem}

\begin{proof}
By exploiting the Pinsker's inequality, we bound the total variation distance as
\begin{align*}
    \text{TV}(p_1,q_1)
    \leq & ~ \sqrt{\frac{1}{2} \text{KL}(p_1,q_1)} \annot{By Pinsker's inequality}\\
    \leq & ~ \sqrt{\frac{1}{2} [A_1 \mathcal{R}(\hat{u}_\theta) + A_2 \mathcal{R}(\hat{u}_\theta)^2]}.
    \annot{By \cref{thm:kl_bound} and let $t_0=0,T=1$} 
\end{align*}
Taking expectation on training data $\{x_i\}_{i=1}^n$, 
\begin{align*}
    \E_{\{x_i\}_{i=1}^n}[ \text{TV}(p_1,q_1) ]
    \lesssim & ~ \E_{\{x_i\}_{i=1}^n}
    [\sqrt{A_1 \mathcal{R}(\hat{u}_\theta) + A_2 \mathcal{R}(\hat{u}_\theta)^2}] \\
    \leq & ~ \E_{\{x_i\}_{i=1}^n}
    [\sqrt{A_1 \mathcal{R}(\hat{u}_\theta)} + \sqrt{A_2 \mathcal{R}(\hat{u}_\theta)^2}] \annot{By $\sqrt{a+b} \leq \sqrt{a} + \sqrt{b}$ for $a,b \geq 0$}\\
    \leq & ~ \sqrt{\E_{\{x_i\}_{i=1}^n}
    [A_1 \mathcal{R}(\hat{u}_\theta)]} +
    \E_{\{x_i\}_{i=1}^n}
    [\sqrt{A_2} \mathcal{R}(\hat{u}_\theta)] 
    \annot{By Cauchy-Schwarz Inequality} \\
    \lesssim & ~ O(  n^{ -\frac{1}{20d} }(\log{n})^{5d_x} ) +
    O(  n^{ -\frac{1}{10d} }(\log{n})^{10d_x} )
    \annot{By \cref{lem:estimation}} \\
    = & ~ O(  n^{ -\frac{1}{20d} }(\log{n})^{5d_x} ).
\end{align*}
This completes the proof.
\end{proof}

\subsection{Proof of \texorpdfstring{ \cref{thm:minimax}}{Theorem \ref{thm:minimax}}}
\label{sec:proof_minimax}

\begin{theorem}
[\cref{thm:minimax} Restated: Nearly Minimax Optimality of Flow Matching Transformers]
\label{thm:minimax_ew}
Let $C$, $C_1$ and $C_2$ be positive constants.
Assume the data distribution satisfies $p_1(x)=\exp(-C_2\norm{x}_2^2/2) \cdot f(x)$,
where $f$ belongs to H\"{o}lder space $f(x) \in \calH^{\beta}(\R^{d_{x}},B)$ (\cref{def:holder_norm_space}) and satisfies 
$ C_1 \geq f(x)\geq C$ for all $x$.
Then, within the Hölder distribution class under the Total Variation (TV) metric, the Flow Matching Transformer achieves the minimax-optimal convergence rate when ${18d}(\beta + 1) =  d_x + 2\beta$.
\end{theorem}

\begin{proof}
Notice that the assumption we made in \cref{thm:minimax} can directly implies \cref{assumption:density_function_assumption_1}.
Then by \cref{thm:dist_esti_tv},
we have the distribution estimation rate in Total Variation distance:
\begin{align*}
    \E_{\{x_i\}_{i=1}^n}[ \text{TV}(p_1,q_1) ]
    =
   O(n^{ -\frac{1}{20d} }(\log{n})^{5d_x}).
\end{align*}

Then, by \cref{lem:minimax_tv},
the distribution rates matches the minimax lower bound up to a $\log{n}$ and Lipschitz constant factors under the setting 
\begin{align*}
20d \beta w=  d_x + 2\beta.
\end{align*}

This completes the proof.
\end{proof}